\documentclass[journal]{IEEEtran}
\usepackage{amssymb}
\usepackage{graphicx}
\usepackage{epstopdf}
\usepackage{soul}
\usepackage{setspace}
\usepackage{algorithmic}
\usepackage{algorithm}
\usepackage{multirow}
\usepackage{color}
\usepackage{soul}
\sethlcolor{yellow}
\usepackage{epstopdf}
\usepackage{textcomp}
\usepackage{amsmath}
\usepackage{amssymb}
\usepackage{wrapfig}
\usepackage{cite}
\usepackage{pdflscape}
\usepackage{array}
\usepackage{longtable,tabu}
\usepackage{arydshln}
\usepackage{url}
\usepackage{amsmath}
\usepackage{amssymb}
\usepackage{amsfonts}
\usepackage{multicol}
\usepackage{booktabs}
\usepackage{url}
\usepackage{tikz}
\usepackage[abs]{overpic}
\usepackage{subfig}
\usepackage{graphicx}
\usepackage{enumitem}
\usepackage[super]{nth}
\usepackage[font={footnotesize}]{caption}
\usepackage[export]{adjustbox}

\newcommand{\bluetext}[1]{{\color{blue}#1}}

\newcolumntype{C}[1]{>{\centering\arraybackslash}m{#1}}

\ifCLASSINFOpdf
\else
\fi
\begin{document}
%
\title{Retinal Microaneurysms Detection using Local Convergence Index Features}
%
%
%


\author{
		\thanks{This work is part of the H\'{e} Programme of Innovation Cooperation, which is financed by the Netherlands Organization for Scientific Research (NWO), dossier No. 629.001.003. 
		\textit{Asterisk indicates corresponding author.}}
	Behdad~Dashtbozorg, 
	Jiong~Zhang, 
	and~Bart~M.~ter~Haar~Romeny,
	\thanks{*B. Dashtbozorg and  J. Zhang are with the Department of Biomedical Engineering, Eindhoven University of Technology, Eindhoven 5600MB, the Netherlands (e-mail: B.Dasht.Bozorg@tue.nl; J.Zhang1@tue.nl).}
	\thanks{B. M. ter Haar Romeny is with the Department of Biomedical and Information Engineering, Northeastern University, Shenyang 110000, China, and also with the Department of Biomedical Engineering, Eindhoven University of Technology, Eindhoven 5600MB, the Netherlands (e-mail: B.M.TerHaarRomeny@tue.nl).}
}

\maketitle

\begin{abstract}
Retinal microaneurysms are the earliest clinical sign of diabetic retinopathy disease. Detection of microaneurysms is crucial for the early diagnosis of diabetic retinopathy and prevention of blindness. In this paper, a novel and reliable method for automatic detection of microaneurysms in retinal images is proposed. In the first stage of the proposed method, several preliminary microaneurysm candidates are extracted using a gradient weighting technique and an iterative thresholding approach. In the next stage, in addition to intensity and shape descriptors, a new set of features based on local convergence index filters is extracted for each candidate. Finally, the collective set of features is fed to a hybrid sampling/boosting classifier to discriminate the MAs from non-MAs candidates. The method is evaluated on images with different resolutions and modalities (RGB and SLO) using five publicly available datasets including the Retinopathy Online Challenge’s dataset. The proposed method achieves an average sensitivity score of 0.471 on the ROC dataset outperforming state-of-the-art approaches in an extensive comparison. The experimental results on the other four datasets demonstrate the effectiveness and robustness of the proposed microaneurysms detection method regardless of different image resolutions and modalities. 
\end{abstract}


%
\IEEEpeerreviewmaketitle

\section{Introduction}\label{sec:introduction}

%
%
%
%
Diabetic retinopathy (DR) is the most common cause of vision loss among people with diabetes and the leading cause of vision impairment and blindness among middle-aged population in the world~\cite{Lee2015}.
Chronically high blood sugar levels from diabetes are associated with increasing damage to the tiny blood vessels in the retina, leading to diabetic retinopathy~\cite{Yau556}. DR can cause vessels in the retina to leak fluid or to bleed, and in advanced stages, new abnormal blood vessels may proliferate on the surface of the retina, which can lead to scarring and cell loss in the retina~\cite{Kohner1999}.
Small swellings in the retina's tiny blood vessels, called microaneurysms (MAs), occur at this earliest stage of the disease~\cite{Kohner1999,Wilkinson20031677}. In digital color fundus images, MAs appear as tiny, reddish isolated dots near tiny blood vessels~\cite{Wilkinson20031677}. 

The detection and analysis of MAs is considered as one of the most important clinical strategies for the early diagnosis of DR and blindness prevention in a cost-effective health care practice. 
However, due to the limited number of ophthalmologists and the large number of people that require screening, an automated computer-aided diagnosis tool can significantly improve the efficiency and reduce the costs in a large-scale screening setting~\cite{dashtbozorg2014automatic}. 

Several methodologies for the detection of MAs have been proposed, most of which were performed in two stages~\cite{Abdelazeem2002,Niemeijer2005,Walter2007555,Cree2008,Mizutani2009,Sanchez2009,Zhang20102237,Ram2011,Giancardo2011,Zhang201278,Fegyver2012,Antal2012,Lazar2013,Zhang2014,Adal20141,Shah2016,Dai2016,Wu2017106,Seoud2016,Wang2017} (see also Table{~\ref{tab:AllDatasetsEvaluation}}). 
In the first stage, several candidates with similar characteristics to MAs are extracted. In the second stage, a set of features is obtained for each candidate and a classification technique is applied for the discrimination of MAs from non-MA candidates.
Niemeijer \textit{et al.}~\cite{Niemeijer2005} proposed a hybrid approach for the candidate extraction which is a combination of mathematical morphology techniques and a pixel classification method. The true MAs are then detected using a k-Nearest Neighbors (kNN) classifier and a set of shape and intensity features.  
For the extraction of initial candidates, Mizutani \textit{et al.}~\cite{Mizutani2009} utilized a modified double-ring filter in which the average pixel intensity value of the inner ring is compared with that of the outer ring, then the candidates are re-examined by removal of blood vessels. The candidates were classified using a three-layered feed-forward neural network.

In the method introduced by Sanchez \textit{et al.}~\cite{Sanchez2009}, a mixture model-based clustering technique is used for the candidate extraction, which is followed by a logistic regression classifier to generate a likelihood for each candidate based on its color, shape, and texture.
Zhang \textit{et al.}~\cite{Zhang20102237} used a multi-scale correlation coefficients-based method and a dynamic thresholding technique for candidates extraction. Then a rule-based classification algorithm was employed for accurate detection of MAs. This approach is further improved by including dictionary learning with a sparse representation classifier~\cite{Zhang201278}.
In the method by Ram \textit{et al.}~\cite{Ram2011} the candidates are selected using a morphology-based approach in which linear structures are extracted in different orientations.  In the next step, a successive rejection-based strategy was designed which passes only true MAs while rejecting false classes of clutter.

Giancardo \textit{et al.}~\cite{Giancardo2011} detected MA candidates using a thresholding technique followed by a Radon transformation at various scanning angles. MAs are then located using a support vector machine (SVM) classifier and features extracted from Radon-space.
The ensemble-based framework proposed by Antal and Hajdu~\cite{Antal2012} detects MAs by selecting the optimal combination of different preprocessing techniques and candidate extractors~\cite{Zhang20102237,Walter2007555,Abdelazeem2002}. 
Lazar and Hajdu~\cite{Lazar2013} proposed a method based on the analysis of locally rotating cross-section profiles, where the local maximum pixels were selected as candidates. This was followed by a peak detection technique applied on 30 produced profiles, and a set of descriptors including the size, height, and shape of the peaks is obtained. The statistical measures of the resulting directional peak descriptors are then used as the set of features for a Bayesian classifier.
Zhang~\cite{Zhang2014} introduced new contextual descriptors which are used in combination with intensity and geometric features in a Random Forest classifier to detect MAs.
Adal \textit{et al.}~\cite{Adal20141} presented a MA candidate selection approach based on scale-invariant interest-points and blob detection theory. They used a local-scale estimation technique and several scale-adapted region descriptors  to characterize detected blob regions. The final classification step was performed using a semi-supervised learning approach.

In the method proposed by Shah \textit{et al.}~\cite{Shah2016}, initial candidates were extracted by removing vessels and local thresholding. A rule-based classifier and a set of statistical features were employed to classify the candidates into MAs and non-MAs.
Dai \textit{et al.}~\cite{Dai2016} extracted MAs candidates by a vessel removal technique and gradient vector analysis. The candidates are then classified using a class-imbalance classifier and several features including geometry, contrast, intensity, edge, texture and region descriptors.
Recently, Seoud \textit{et al.}~\cite{Seoud2016} introduced a new set of shape features called Dynamic Shape Features which are employed for the detection of red lesions in retinal images.
Wu \textit{et al.}~\cite{Wu2017106} presented a candidate selection approach similar to Lazar’s method~\cite{Lazar2013} using profile analysis and region growing. For the final classification stage, the authors used several local and profile-based features and different classifiers such as KNN and Adaboost.
In the method proposed by Wang \textit{et al.}~\cite{Wang2017} candidates were located using a dark object filtering process. Afterwards, singular spectrum analysis is employed to decompose cross-section profiles of extracted objects and reconstruct a new one. A kNN classifier and a set of statistical features of profiles were used to discriminate the MAs from non-MAs candidates. 

In addition to the two-stage approaches, Pereira \textit{et al.}~\cite{Pereira2014179} proposed a multi-agent system method for MAs segmentation using gradient patterns and Gaussian fitting parameters in different directions. 
Quellec \textit{et al.}~\cite{Quellec2008} modeled the MAs with 2-D rotation-symmetric generalized Gaussian functions and used a supervised template matching technique in wavelet-subbands for the MAs detection. The optimal adapted wavelet transform for MA detection was found by applying the lifting scheme framework and then MAs were detected using template matching in the wavelet domain.
{Although in  recent years deep neural networks have gained popularity for the DR detection and in general in the field of computer vision, only a few papers report specifically on the detection of MAs using deep learning techniques~\cite{haloi2015improved,Shan2016}. 
The results of MA candidate extraction and MA detection algorithms presented in the literature are listed and compared in Tables~\ref{tab:CandidateEvaluation} and~\ref{tab:AllDatasetsEvaluation}.}
%
%
%

Despite the many published approaches, described above~\cite{Abdelazeem2002,Niemeijer2005,Walter2007555,Cree2008,Quellec2008,Mizutani2009,Sanchez2009,Zhang20102237,Ram2011,Giancardo2011,Zhang201278,Fegyver2012,Antal2012,Lazar2013,Pereira2014179,Zhang2014,Adal20141,Shah2016,Dai2016,Wu2017106,Seoud2016,Wang2017,haloi2015improved,Shan2016}, accurate detection of MAs is still a challenging task. The detection of MAs depends on the image properties and the characteristics of the imaging device such as resolution, image modality, inter/intra-image illumination and contrast variation and compression technique. On the other hand, the variations in MA size, shape and proximity to vessels can also increase the difficulty of identifying them in retinal images.
In this paper, we propose a novel method for the detection of MAs using local convergence index filters (LCF) and a random undersampling boosting classifier (RUSBoost). 
The method extracts several candidates and generates a set of features for each candidate depending upon their intensity, shape and LCF responses. The true MA candidates are then selected using a hybrid sampling/boosting classifier to avoid the drawback of imbalanced data learning and to improve the performance of MA detection.
The major contributions of this paper can be summarized as follows.\\
\vspace{-1.5em}
\begin{enumerate}[label=(\roman*)]
	\item A novel method is proposed for accurate and reliable detection of microaneurysms with the possibility of applying this method in large screening setups. The method outperforms state-of-the-art techniques.
	\item A new MA candidate extraction technique is employed which extracts the potential MA candidates with a notable improvement in sensitivity;
	\item A new set of features based on local convergence index filters is introduced which are then incorporated into a hybrid sampling/boosting classifier;
	\item The proposed approach is evaluated on five datasets with two image modalities. The high performance of our method provides an indication of the detection power of the proposed approach in handling difficult cases such as subtle MAs and the ones close to vessels;
	\item We published two new publicly available retinal image databases, RC-RGB-MA (RGB fundus camera) and RC-SLO-MA (scanning laser ophthalmoscope) to evaluate microaneurysms detection methods on different image modalities.  Moreover, we provided a software tool for microaneurysm annotation (RC-MAT) helping the experts to collect more labeled MAs. The tool is publicly available as a software package;

\end{enumerate}

This paper is organized as follows. In Section~\ref{sec:methodology}, the proposed method for microaneurysms detection is
presented. In Section~\ref{sec:Validation}, we show the experimental results with performance evaluations. Finally, we discuss and conclude our method in Section~\ref{sec:Discussion}.
\vspace{-1em}

\begin{figure}[!t]
	\centering
	\includegraphics[trim={0.5cm 0.5cm 0.5cm 0.5cm},clip,width=0.5\textwidth]{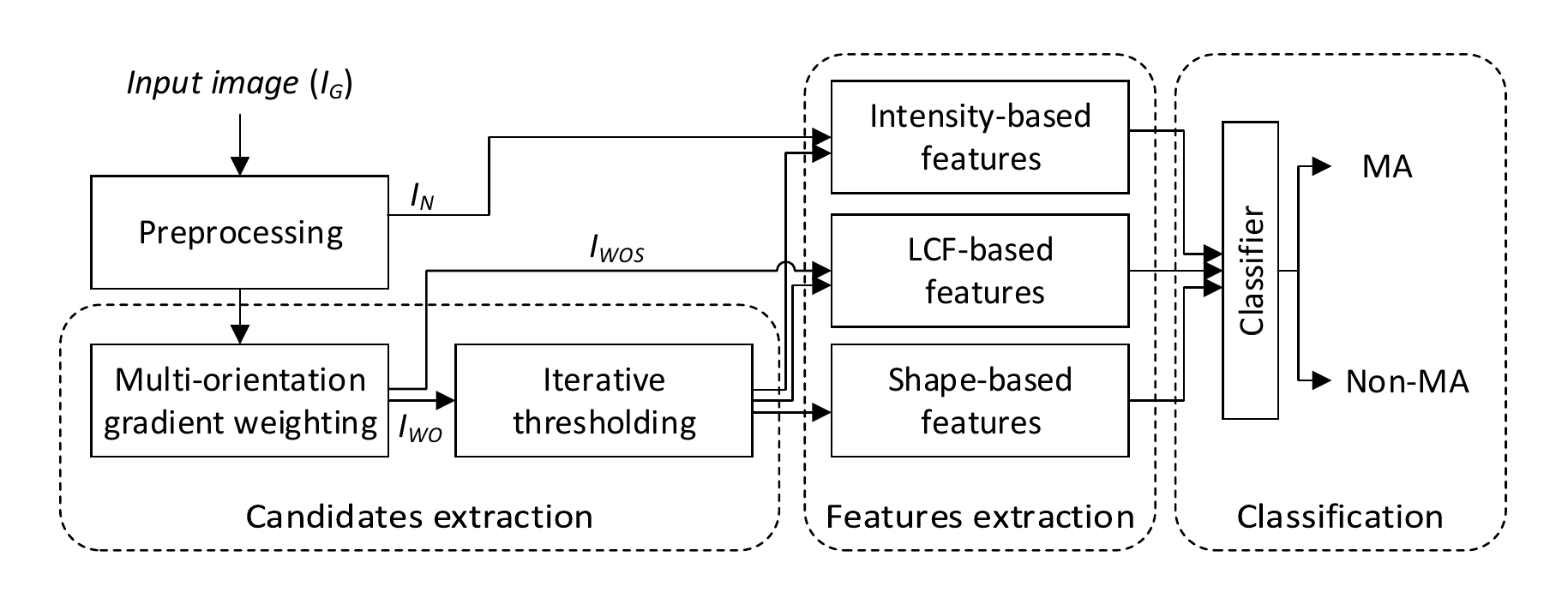}%
	\caption{Block diagram for the proposed microaneurysm detection method.}
	\label{fig:preprocessing:blockdiagram}
	\vspace{-1.5em}
\end{figure}

\section{Methodology}\label{sec:methodology}

\figurename{~\ref{fig:preprocessing:blockdiagram}} shows the pipeline of the proposed algorithm. After the preprocessing step, the main phases are 1) candidates extraction using multi-scale multi-orientation gradient weighting and iterative thresholding; 2) features extraction using local convergence index filters; and 3) microaneurysm detection by training a RUSBoost classifier.
\subsection{Preprocessing}\label{sec:Preprocessing}

\begin{figure}[!t]
	\centering
	\subfloat{}{\includegraphics[trim={2.5cm 0.5cm 2.5cm 0.5cm},clip,width=0.23\textwidth]{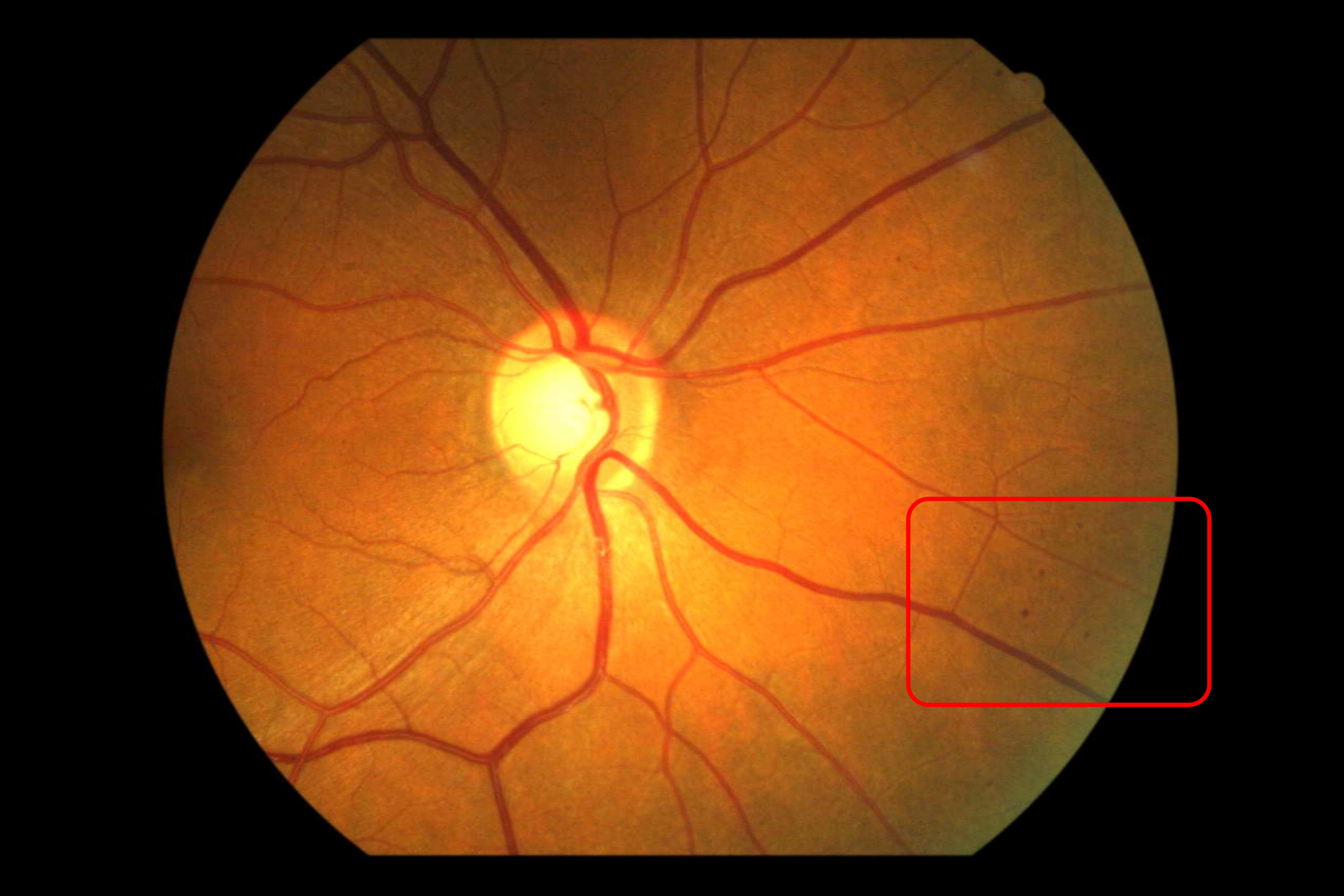}%
		\label{fig:preprocessing:original}}
	\hfil
	\subfloat{}{\includegraphics[trim={23.5cm 4.95cm 3.65cm 12.7cm},clip,width=0.23\textwidth,cfbox=red 1pt 0pt]{OriginalImageBrightRect-eps-converted-to.pdf}%
		\label{fig:preprocessing:originalCrop}}

	\subfloat{}{\includegraphics[trim={2.5cm 0.5cm 2.5cm 0.5cm},clip,width=0.23\textwidth]{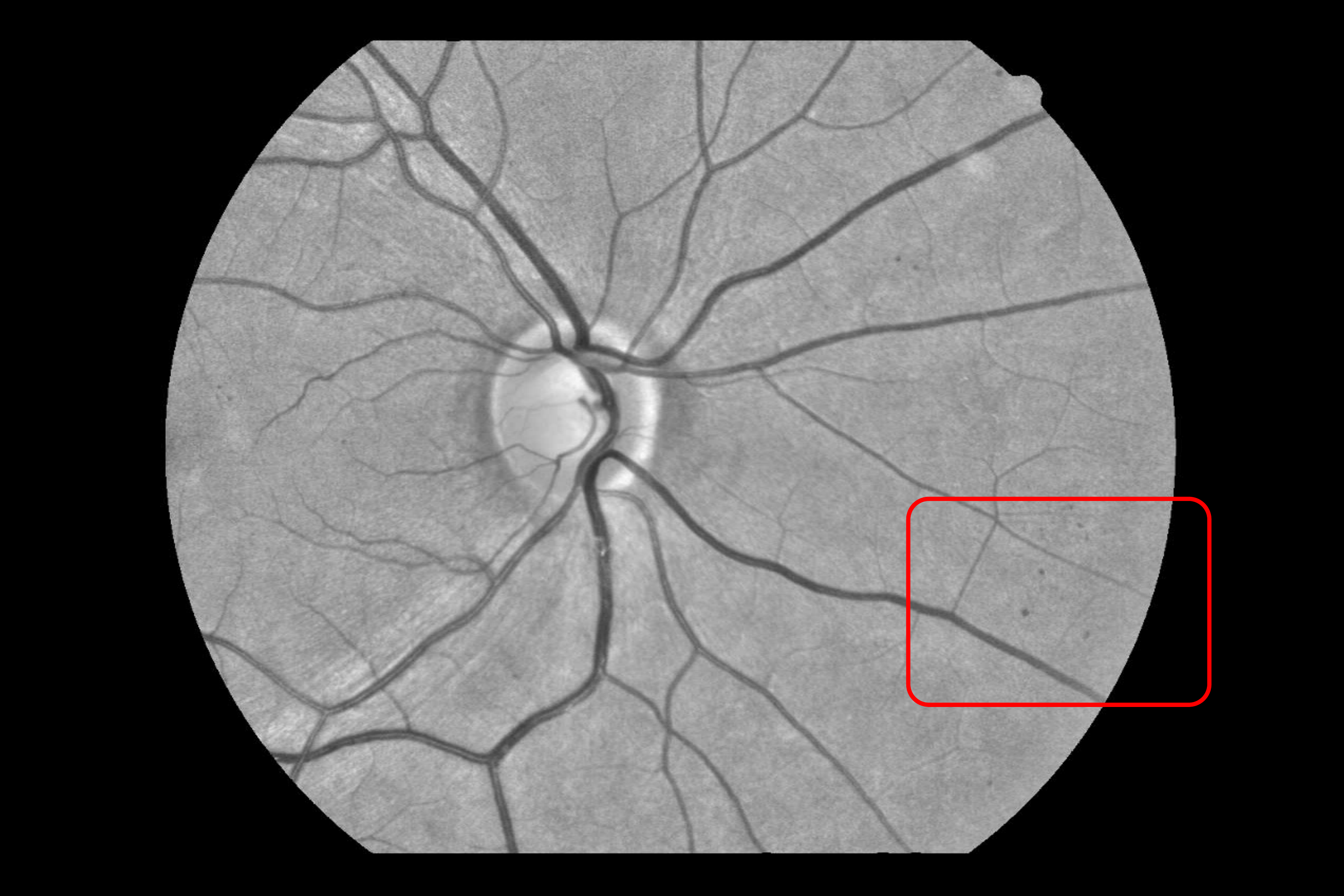}%
		\label{fig:preprocessing:normalized}}
	\hfil
	\subfloat{}{\includegraphics[trim={23.5cm 4.95cm 3.65cm 12.7cm},clip,width=0.23\textwidth,cfbox=red 1pt 0pt]{NormalizedImage2BrightRect-eps-converted-to.pdf}%
		\label{fig:preprocessing:normalizedCrop}}
	
	\caption{Preprocessing results on original size image and small patch; \nth{1} row: Original color fundus image $I$; \nth{2} row: Normalized green channel $I_N$. The patch size is $550 \times 350\; px$.}
	\label{fig:preprocessing}
	\vspace{-1.5em}
\end{figure}

Since the green channel of retinal images in RGB datasets provides a better contrast between microaneurysm and  background, we only use the green channel ($I_G$) in our experiments.
As a result of the acquisition process, very often the retinal images are non-uniformly illuminated and exhibit local luminosity and contrast variability. In order to make the MB detection more robust, each image is preprocessed using the method proposed by Foracchia \textit{et al.}~\cite{foracchia2005luminosity}, which normalizes both luminosity and contrast based on a model of the observed image. 
The local normalization reduces luminosity and contrast variation of retinal images, and it improves the visibility of the lesions. The result of preprocessing on a sample retinal image is demonstrated in \figurename{~\ref{fig:preprocessing}}.

\subsection{Gradient Weighting}

We compute multi-scale multi-orientation weights for each pixel in a normalized image $I_N$ based on the gradient magnitude at that pixel.  The gradient magnitude values for different scales ($\sigma_G$) and different orientation ($\theta_G$) are obtained by the convolution of the normalized image with the rotated first order derivative of the Gaussian kernel:

\begin{figure}[!t]
	\centering
	\includegraphics[trim={3.5cm 0cm 12cm 1cm},clip,width=0.48\textwidth]{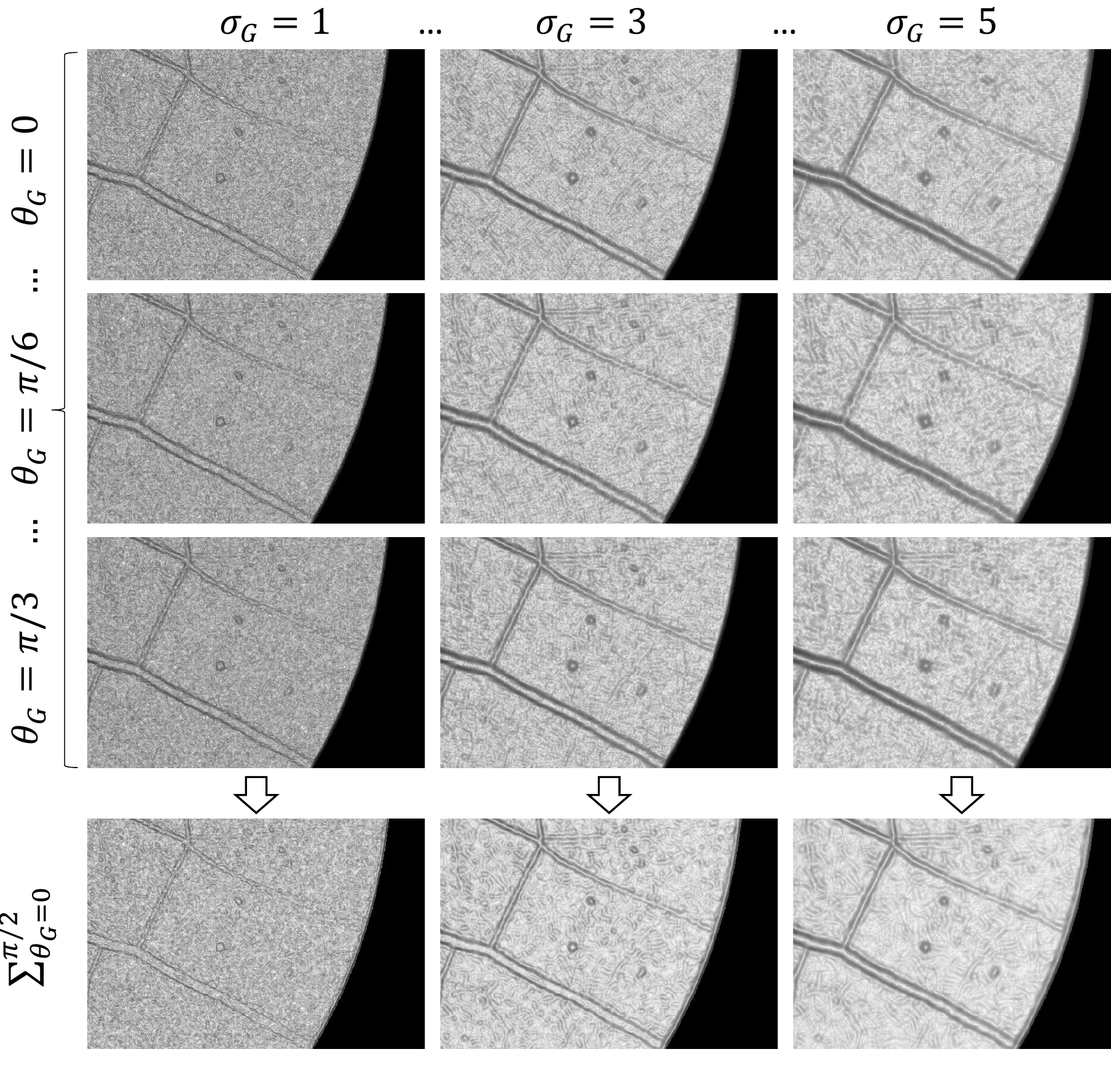}%
	\vspace{-0.5em}	
	\caption{Gradient-weighted image patches in different scales and orientations. The last row shows the summation results over all orientations.}
	\label{fig:multiscalemultiorientation}
	\vspace{-1.5em}
\end{figure}


\begin{figure}[!t]
	
	\centering
	\subfloat{}{\includegraphics[trim={4cm 2cm 3.7cm 2cm},clip,width=0.23\textwidth]{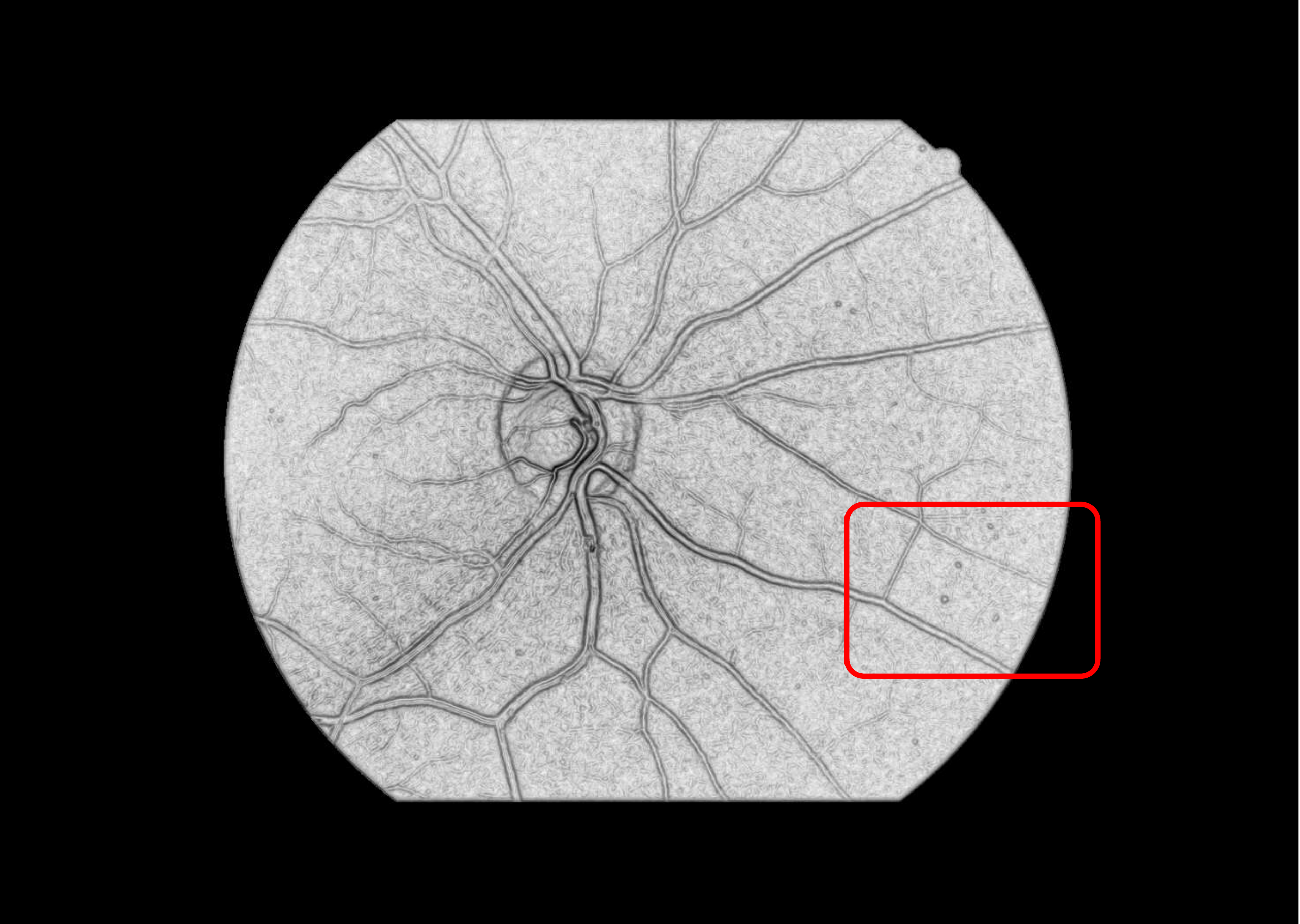}%
		\label{fig:GWimageWhole}}
	\hfill
	\subfloat{}{\includegraphics[trim={17cm 4.7cm 4cm 9.85cm},clip,width=0.23\textwidth,cfbox=red 1pt 0pt]{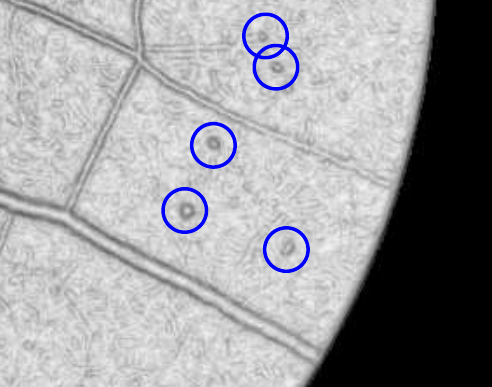}%
		\label{fig:GWImageCrop}}
	\caption{Multi-scale and multi-orientation gradient-weighted image $I_{WOS}$ where the donut-shaped MBs are annotated by blue circles.}
	\label{fig:GWImage}
	\vspace{-1.5em}
\end{figure}

\begin{equation}
	\begin{split}
		I_{M}(x,y,\sigma_G,&\theta_G) = \bigg{[} \left( \frac{\partial G(x,y,\sigma_G)}{\partial x}\bigg{|}_{\theta_G} \otimes I_{N}(x,y)\right )^{2} 
		\\
		&+\left( \frac{\partial G(x,y,\sigma_G)}{\partial y}\bigg{|}_{\theta_G} \otimes I_{N}(x,y)\right )^{2} \bigg{]}^{1/2}
	\end{split}
	\label{eq:GaussianMag}
\end{equation}
where $\theta_G\in\{0,\pi/12,...\pi/2\}$ denotes the rotation angle of the Gaussian derivative kernel and the Gaussian kernel is given by
$G(x,y,\sigma_G) = ({1}/{\sqrt{2\pi\sigma_{G}^2}})e^{({-{x^2+y^2})/{2\sigma_{G}^2}}}$
where $\sigma_{G}\in\{1,2,3,4,5\} px$ is the standard deviation of the Gaussian kernel which is used as scaling parameter.

The weight of a pixel is inversely related to the gradient values at the  pixel location, so that for pixels with small gradient magnitude (smooth regions), the weight is large, and for pixels with large gradient magnitude (such as on the edges), the output weight is small. The gradient-weighted image is obtained by
\begin{equation}
	I_{W}(x,y,\sigma_G,\theta_G) =\frac{1-I_{M}(x,y,\sigma_G,\theta_G)^{2}}{1+I_{M}(x,y,\sigma_G,\theta_G)^{2}}
	\label{eq:gradient-weights}
\end{equation}
Samples of gradient-weighted images of a small patch for different scales and orientations are shown in \figurename{~\ref{fig:multiscalemultiorientation}}.
   


The final multi-orientation gradient-weighted image $I_{WO}$ is defined as the sum of the individual gradient-weighted images in all orientations for each scale separately via
\begin{equation}
	I_{WO}(x,y,\sigma_G) =\sum_{\theta_{G}=0}^{\pi/2}I_{W}(x,y,\sigma,\theta_G)
	\label{eq:GW-averageOrientation}
\end{equation}
The multi-orientation gradient-weighted images are used in the successive thresholding process to extract candidates. Examples of the obtained results are illustrated in the last row of 
\figurename{~\ref{fig:multiscalemultiorientation}}.
To cover all microaneurysms’ sizes, the multi-orientation gradient-weighted images are obtained at several scales. The final multi-orientation and multi-scale gradient-weighted image $I_{WOS}$ is obtained by the summation of $I_{WO}$ over all selected scales
\begin{equation}
	I_{WOS}(x,y) =\sum_{\sigma_{G}=1}^{5}I_{WO}(x,y,\sigma_G) 
	\label{eq:GW-averageScale}
\end{equation}

\figurename{\ref{fig:GWImage}} shows the final gradient-weighted image $I_{WOS}$, where the MAs appeared as hollow circles (like donuts).


\subsection{MA Candidate Extraction}\label{sec:CandidateExtraction}
The candidate extraction step is one of the main phases in automated detection of microaneurysms. In this subsection, a suitable candidate selection algorithm is proposed which decreases the false positive rate as well as the complexity and the computation time by reducing the number of objects for further analysis.


In this step, an iterative thresholding process is applied on the multi-orientation gradient-weighted images ($I_{WO}$) to obtain a set of binary images $I^{t}_{\sigma_{G}}$ corresponding to different threshold values ($t$) in each scale separately.  Using the connected component analysis in each binary image, the objects ($C^{t}_{k}$) with a hole inside satisfying the area, eccentricity and extent constraints are identified and selected as MA candidates. In our experiments, if the object's area, eccentricity and extent are smaller than $300\;px$, 0.9 and 0.3, respectively, then the object is included in the set of selected candidates $\mathbb{M}^{t}_{\sigma_{G}}$. The parameters are set to the values which produce the highest possible sensitivity.

\begin{figure}[!t]
	\centering
	\subfloat{}{\includegraphics[trim={3cm 0.4cm 3cm 0.4cm},clip,width=0.23\textwidth]{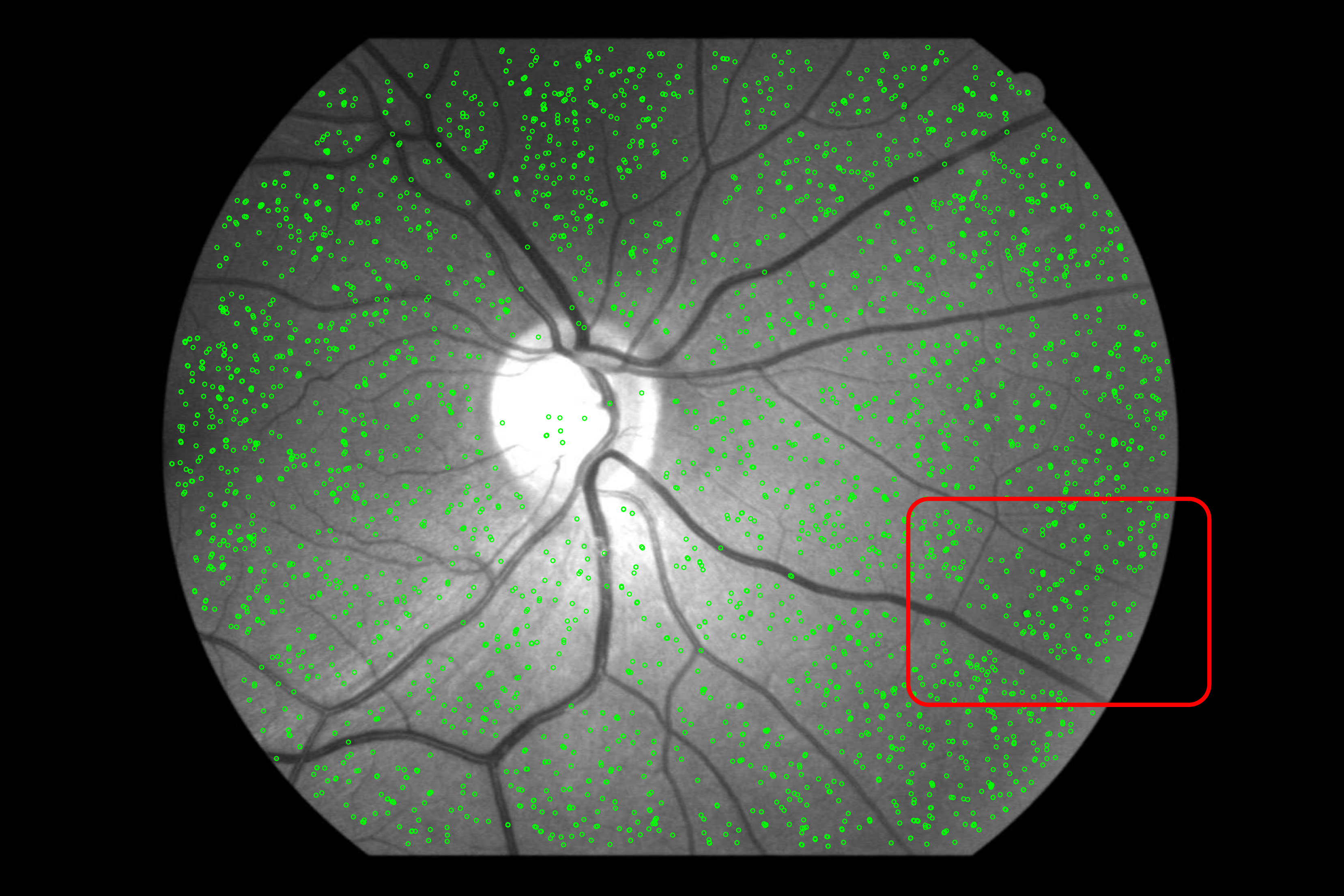}%
		\label{fig:CandidateSelectionOriginal}}
	\hfill
	\subfloat{}{\includegraphics[trim={22.8cm 4.4cm 3.5cm 12.3cm},clip,width=0.23\textwidth,cfbox=red 1pt 0pt]{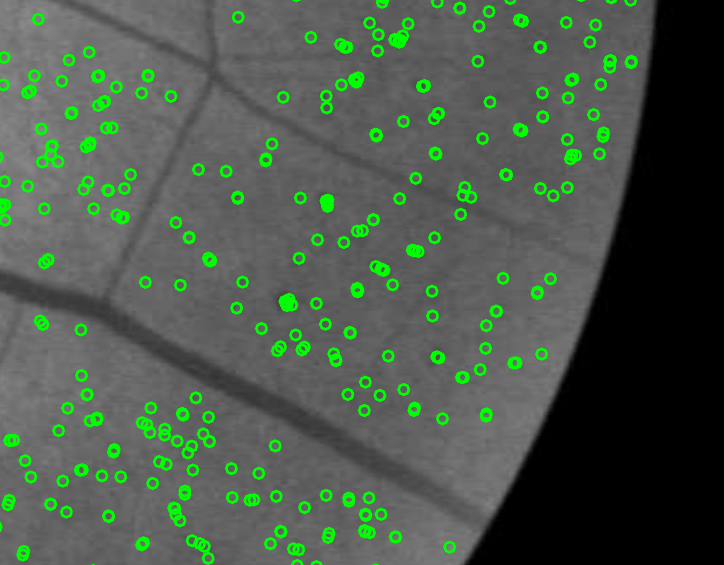}
		\label{fig:CandidateSelectionCrop}}
	
	\caption{Candidates extraction results where the MA candidates are indicated by green circles.}
	\label{fig:CandidateSelection}
	\vspace{-1.5em}
\end{figure}

The final set of MA candidates, $\mathbb{M}$, is the union of all selected candidates at different scales and threshold. The detailed process of the MA candidates extraction is given in Algorithm~\ref{alg:thresholding}. 
It is worth noting that the proposed candidate extraction technique excludes the points on the vessels, which reduces the detection error in the final classification step. \figurename{~\ref{fig:CandidateSelection}} shows examples of the final extracted candidates.

\begin{algorithm}[!h]
	\caption{Candidates extraction}
	\label{alg:thresholding}
	{\footnotesize 	
		\begin{algorithmic}
			\STATE \textbf{Input:} $I_{WO}(x,y,\sigma_G)$
			\STATE \textbf{Output:} Set of microaneurysms candidates $\mathbb{M}$
			\FOR{$\sigma_G = 1$ to $5$ \textbf{step} 1} 
				\FOR{$t=0.1$ to $1$ \textbf{step} 0.05}
				
					\STATE $I^{t}_{\sigma_G}(x,y)=\left\{\begin{matrix}
					1  & \text{if } I_{WO}(x,y,\sigma_G)<t \\ 
					0 &\text{otherwise} 
					\end{matrix}\right.$
					\STATE $C^{t}_{k} \leftarrow $ connected components in binary image $I^{t}_{\sigma_G}$ 
					\STATE where $k\in\{1,....,N_t\}$ and  $N_t$ is total number of components
					\FOR {$k = 1$ to $N_t$}
						\IF {$\textrm{Area}(C^{t}_{k})<300	\wedge	 
							\textrm{Eccentricity}(C^{t}_{k})<0.9	\wedge	 
							\textrm{EulerNumber}(C^{t}_{k})\leq 0	\wedge	 
							\textrm{Extent}(C^{t}_{k})<0.3	$}
							\STATE $\displaystyle \mathbb{M}_{\sigma_G}^{t} \leftarrow \mathbb{M}_{\sigma_G}^{t} \cup C^{t}_{k}$
						\ENDIF
					\ENDFOR
				\ENDFOR	
				\STATE $\displaystyle \mathbb{M}_{\sigma_G} \leftarrow \bigcup_{t}^{ }\mathbb{M}_{\sigma_G}^{t} $		
			\ENDFOR			
			\STATE $\displaystyle  \mathbb{M} \leftarrow \bigcup_{\sigma_{G}}^{}\mathbb{M}_{\sigma_G} $
		\end{algorithmic}
	}

\end{algorithm}
\vspace*{-1.5em}

%

\begin{table*}[!t]
	\renewcommand{\arraystretch}{0.9}
	\centering
	\caption{Description of features for the MA classification.}
	\vspace{-0.5em}
	{\footnotesize 	
		\begin{tabular}{p{.7cm}p{3.4cm}p{13cm}}
			\toprule
			Notation & Name& Description\\			
			\midrule
			\midrule
			
			$f_{1}$			&$G{c}$ 									&Green intensity value at the center of candidate region\\
			$f_{2-4}$		&$G_{mean}, G_{max}, G_{min}$				&Mean, maximum and minimum green intensity values of candidate region\\
			$f_{5-7}$		&$G_{Nmean}, G_{Nmax}, G_{Nmin}$			&Mean, maximum and minimum green intensity values of candidate neighborhood region\\
			$f_{8-9}$		&$S_{Area}, S_{ConA}$ 						&Area and convex area of candidate region\\
			$f_{10}$		&$S_{Sol}$ 									&Ratio of the area to the convex area\\
			$f_{11}$		&$S_{Ext}$ 									&Ratio of pixels in the candidate region to pixels in the total bounding box\\
			$f_{12}$		&$S_{Per}$									&Distance around the boundary of the region\\
			$f_{13}$		&$S_{CirD}$ 								&Diameter of a circle with the same area as the region\\
			$f_{14-15}$		&$S_{AxiA},S_{AxiB}$ 						&Major and minor axes lengths of the ellipse with a same normalized \nth{2} central moments as the region\\
			$f_{16}$		&$S_{Ecc}$									&Ratio of distance between the foci and the major axis length of the ellipse with a same \nth{2} moment as the region\\
			$f_{17}$		&$S_{Eul}$ 									&Number of objects in the region minus the number of holes in those objects\\
			$f_{18-20}$		&$F_{NARF}, F_{NSBF}, F_{NSEF}$ 				&ARF, SBF and SEF filters responses on normalized image $I_{N}$\\
			$f_{21-23}$		&$R_{NARF}, R_{NSBF}, R_{GSEF}$ 				&Estimated radii using ARF, SBF and SEF filters on image $I_{N}$\\
			$f_{24-26}$		&$F_{WARF}, F_{WSBF}, F_{WSEF}$ 				&ARF, SBF and SEF filters responses on multi-scale multi-orientation gradient-weighted image $I_{WOS}$\\
			$f_{27-29}$		&$R_{WARF}, R_{WSBF}, R_{WSEF}$ 				&Estimated radii using ARF, SBF and SEF filters on image $I_{WOS}$\\
			\bottomrule
			\noalign{\vskip 1mm}
			\multicolumn{3}{l}{ARF: adaptive ring filter; SBF: sliding band filter; SEF: super-elliptical filter. }
		\end{tabular}
	}
	\label{tab:featuresList}%
	\vspace{-1.5em}
\end{table*}%

\subsection{Feature Extraction}\label{sec:FeatureExtraction}
Extracting suitable features and descriptors for the candidate regions is an important step for the final classification stage. Since the MAs appear in different colors and sizes, several shape and intensity features are extracted. The feature set is completed by including the responses and the estimated radii of different local convergence index filters (LCF).
The rest of this subsection describes the 29 proposed features which we have defined to characterize and classify MAs.
Table{~\ref{tab:featuresList}} contains the description of the features which we use in the proposed system.


\subsubsection{Intensity-based features}\label{sec:IntensityBasedFeatures}

These features are descriptors indicating the darkness of MAs compared to their neighborhood background.    
The intensity features are extracted at the center of the candidate, inside the whole candidate object and also in a square neighborhood region which is 3 times as large as the candidate area. The neighborhood region is centered on the center of candidate. 
As shown in Table~\ref{tab:featuresList} for every candidate in the $\mathbb{M}$ set, the average, maximum and minimum of the green intensity values are obtained for the candidate and neighborhood regions, separately. 

%

\subsubsection{Shape-based features}\label{sec:ShapeBasedFeatures}

\begin{figure}[!t]
	\centering

	\subfloat[]{\includegraphics[trim={22cm 10cm 22.5cm 10cm},clip,width=0.155\textwidth]{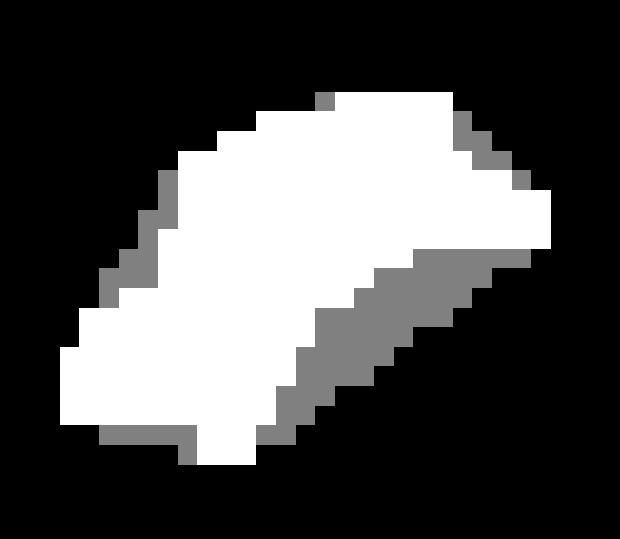}%
		\label{fig:Features:Shape:2}}
	\hfill
	\subfloat[]{\includegraphics[trim={22cm 10cm 22.5cm 10cm},clip,width=0.155\textwidth]{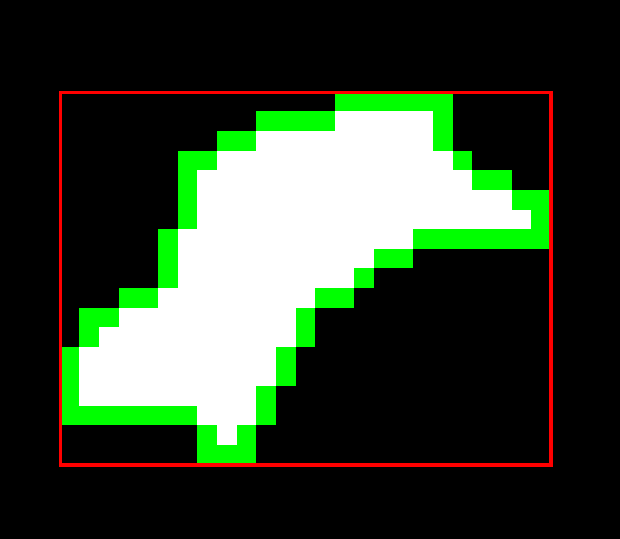}%
		\label{fig:Features:Shape:3}}
	\hfill
	\subfloat[]{\includegraphics[trim={22cm 10cm 22.5cm 10cm},clip,width=0.155\textwidth]{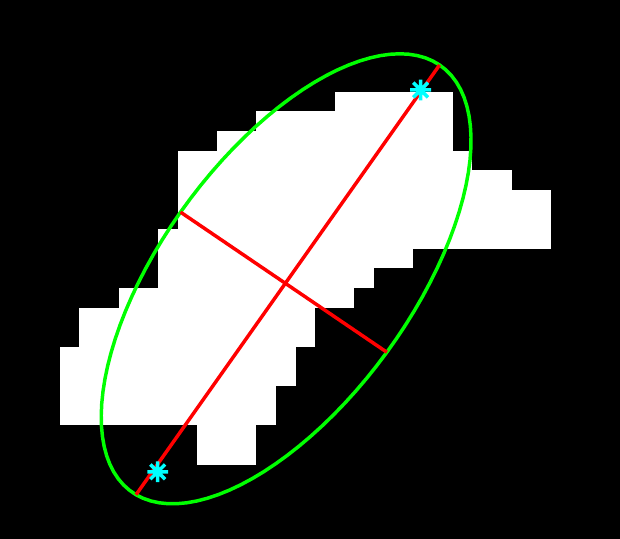}%
		\label{fig:Features:Shape:6}}
	\vspace{-0.5em}
	\caption{Shape features representation (a) candidate competent in white and convex hull region in gray; (b) bounding box in red and perimeter in green; (c) ellipse that has same second moments as the region (in green) with minor and major axes (in red) and ellipse's foci are shown with blue stars.}
	\label{fig:Features:Shape}
	\vspace{-1.5em}
\end{figure}

Since MAs are small and they appear as round structures with a diameter less than $125 \mu m$, the following shape-based features are extracted for each candidate region:

\begin{itemize}
\setlength\itemsep{-1em}
	\item Area ($S_{Area}$): area of candidate region specified by the actual number of pixels (white pixels in \figurename{~\ref{fig:Features:Shape:2}}).\\
	\item Convex area ($S_{ConA}$):  area of candidate convex region specified by the actual number of pixels (white and gray pixels in \figurename{~\ref{fig:Features:Shape:2}}).\\
	\item Solidity ($S_{Sol}$): ratio of the area of candidate ($S_{Area}$) over the convex area ($S_{ConA}$).\\
	\item Extent ($S_{Ext}$): ratio of $S_{Area}$ to the pixels in the bounding box as shown by red color in \figurename{~\ref{fig:Features:Shape:3}}.\\
	\item Perimeter ($S_{Per}$): distance around the boundary of the region by calculating the distance between each adjoining pair of pixels.\\
	\item Circularity ($S_{CirD}$): diameter of a circle with the same area as the region which is equal to $\sqrt{4S_{Area}/\pi}$. \\
	\item Ellipticity ($S_{AxiA},S_{AxiB}$): lengths of the major and minor axes of the ellipse that has the same normalized second central moments as the candidate region. The major and minor axes are depicted by red lines in \figurename{~\ref{fig:Features:Shape:6}}.\\
	\item Eccentricity ($S_{Ecc}$): ratio of distance between the foci (blue stars in  \figurename{~\ref{fig:Features:Shape:6}}) and the major axis length ($S_{AxiA}$) of the ellipse with a same \nth{2} moment as the region.\\
	\item Euler number ($S_{Eul}$):   number of objects in the region minus the number of holes in those objects. 
\end{itemize}

\subsubsection{LCF-based features}\label{sec:LCFBasedFeatures}

The LCF filters are based on gradient convergence and not intensity and as such can detect low contrast MAs which otherwise would be easily lost in the background noise. Additionally, the convergence evaluation in a regional band allows the reduction of uncertainty caused by noise. We first give explanations about the local convergence index filters and then introduce the set of LCF-based features extracted for each candidate.
The convergence index (CI) filters are suitable for the detection of convex shapes and objects with a limited range of sizes regardless of their contrast with respect to the background.
The CI filters evaluate the convergence degree of gradient vectors within a local area (support region) towards a pixel of interest~\cite{kobatake1999convergence}. 

Given an input image $I(x,y)$, for each pixel with spatial coordinates ($x,y$), the convergence index (CI) is defined by

\begin{equation}
CI(x,y)=\frac{1}{M}\sum_{(\theta_{i},m)\in S}\cos \left (\varphi \left (x,y,\theta_{i},m\right) \right),
\label{eq1}
\end{equation}
where $M$ is the number of points in the filter support region $S$, and $\varphi \left (x,y,\theta_{i},m\right)$ is the orientation angle of the gradient vector at the polar coordinate  ($\theta_{i},m$) with respect to the line, with direction $i$, that connects ($\theta_{i},m$) to ($x,y$). The angular difference $\varphi$, is given by 

\begin{equation}
\begin{split}
&\varphi  (x,y,\theta_{i},m) = \theta_{i}-\alpha(x,y,\theta_{i},m),
\\ 
&\alpha\left (x,y,\theta_{i},m \right ) 
\\ &= \tan^{-1} \left ( \frac{\frac{\partial}{\partial x}I(x+m\times \sin (\theta_{i}),y+m\times \cos (\theta_{i}))}{\frac{\partial}{\partial y} I(x+m\times \sin (\theta_{i}),y+m\times \cos (\theta_{i}))} \right ),
\end{split}
\end{equation}
where  $\alpha$ is the image gradient orientation within the convergence filter support region.
The support region polar coordinates are denoted by the radial coordinate $m$, the distance from point of interest ($x,y$) in pixel, and angular coordinate $\theta_{i}$ which is sampled with $N$ equally spaced radial lines ($\theta_{i} = \frac{2\pi}{N}(i-1),\;i \in \{1,...,N \}$). The set of radial lines is emerging from the point where the filter is being applied to, and is equally distributed over a circular region centered at the point of interest ($x,y$).

\begin{figure}[!t]
	\centering
	\subfloat[]{\includegraphics[width=1.11in]{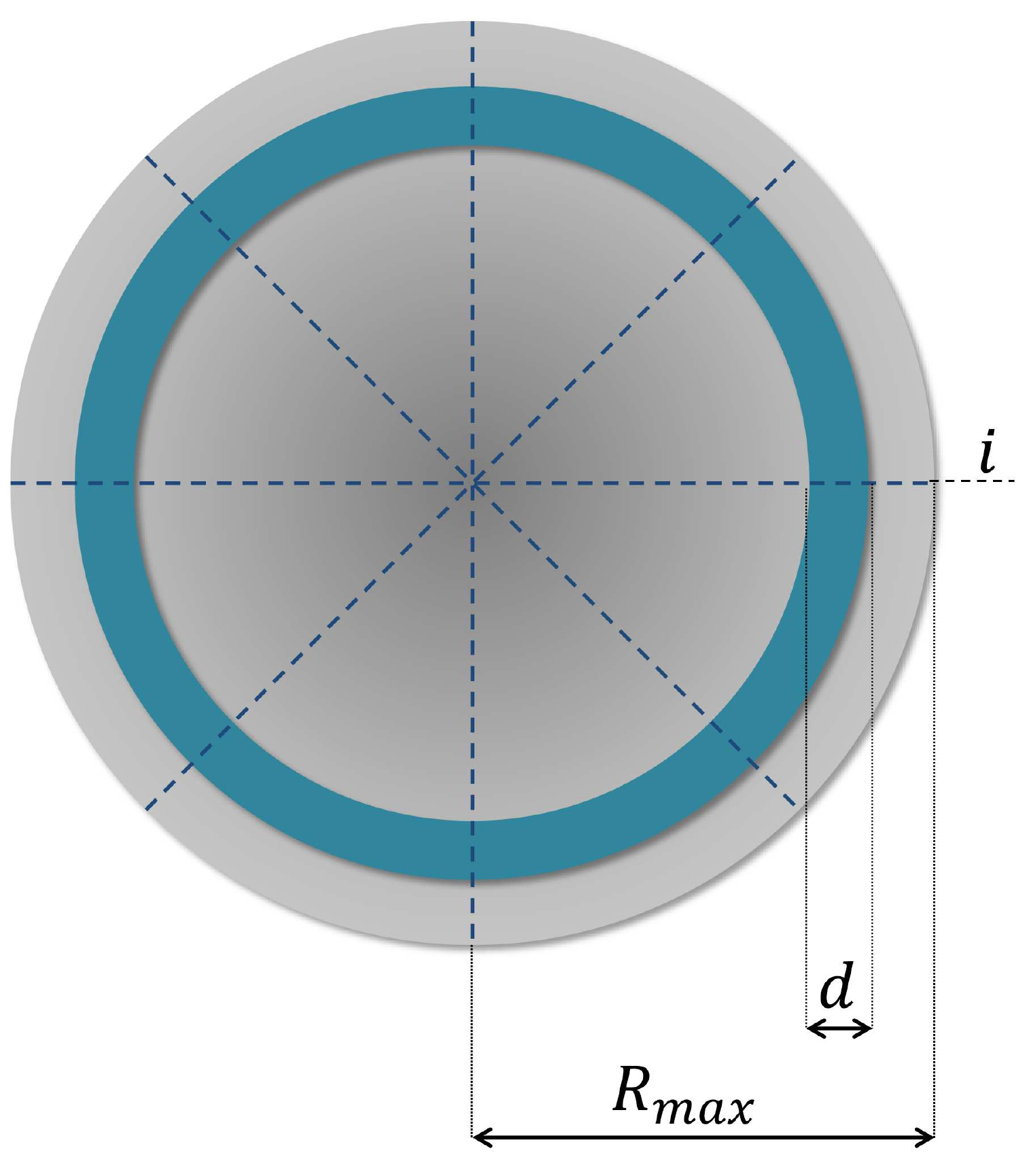}%
		\label{fig:ARF}}
	\hfil
	\subfloat[]{\includegraphics[width=1.11in]{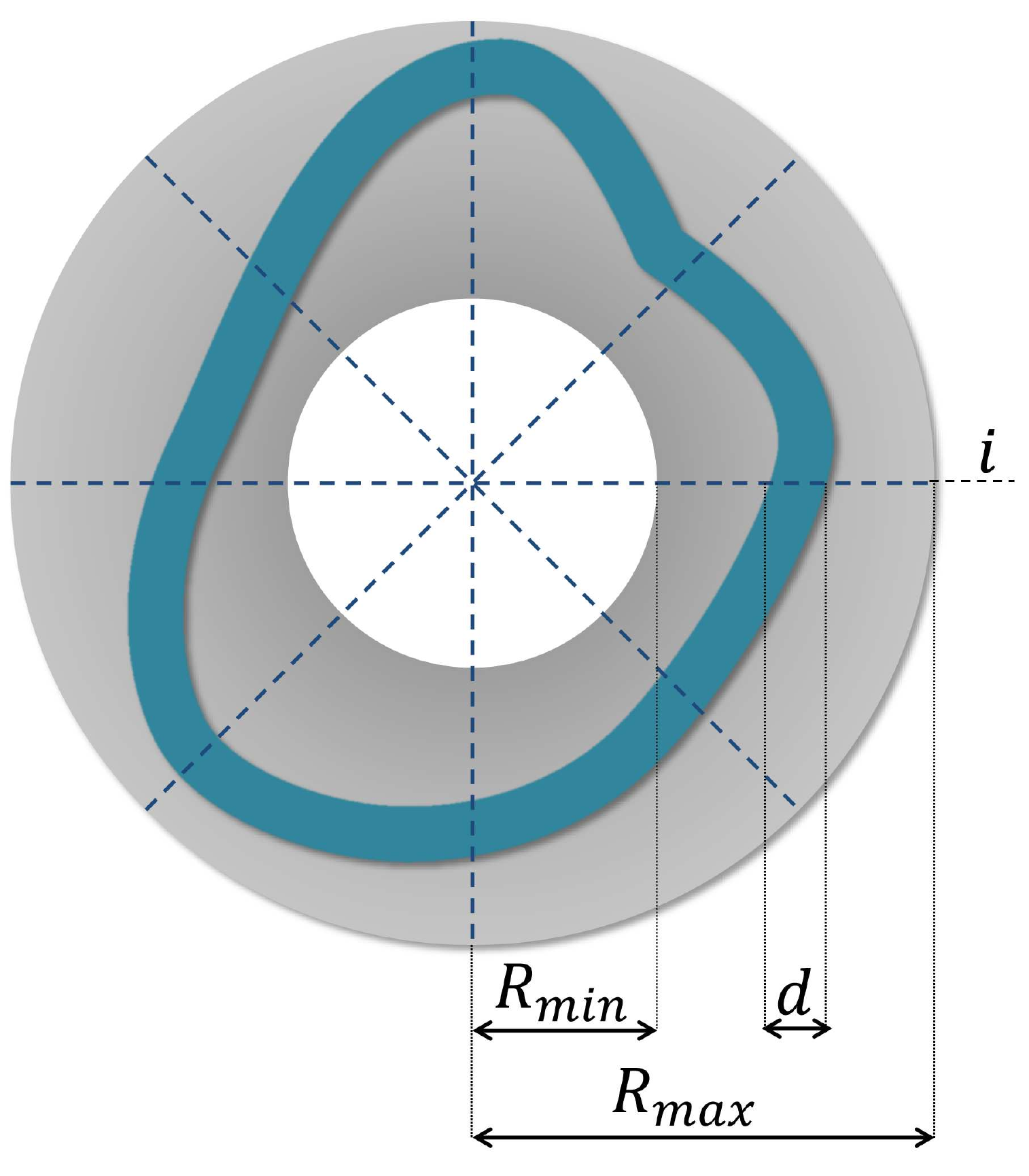}%
		\label{fig:SBF}}
	\hfil
	\subfloat[]{\includegraphics[width=1.11in]{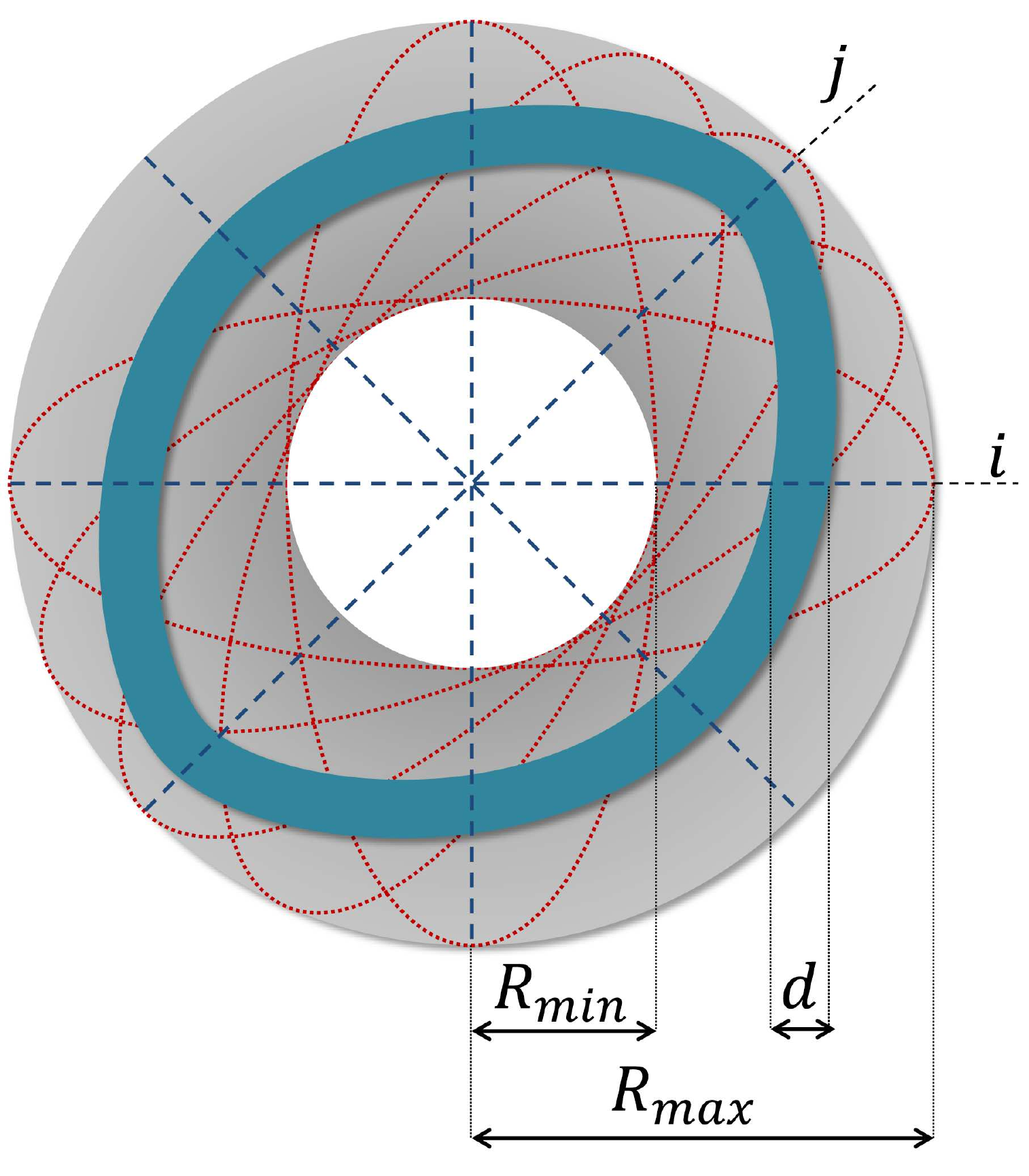}%
		\label{fig:SEF}}
	\vspace{-0.5em}
	\caption{Schematics of  (a) adaptive ring filter, (b) sliding band filter and (c) super-elliptical filter, where the support region lines are depicted with dashed lines and the support region is specified in blue.}
	\label{fig:LCFs}

	\vspace{-1.5em}
\end{figure}

Several CI filters have been proposed according to the way how the support region  is defined~\cite{kobatake1999convergence}. 
Among different CI filters,  the adaptive ring filter (ARF)~\cite{wei1999detection}, the sliding band filter (SBF)~\cite{pereira2007detection} and super-elliptical filter (SEF)~\cite{dashtbozorg2016SEF} are more suitable for the MA detection, since they can be parameterized to use a narrow band for spotting the donut-shaped objects in the gradient-weighted images.

As shown in \figurename{~\ref{fig:ARF}}, the ARF has a ring-shaped region of support and its radius changes adaptively. The response of the ARF is obtained via:
\begin{equation}
F_{ARF}(x,y)=  \max\limits_{0\leq r\leq R_{max}} \frac{1}{N\times d} \sum_{i=1}^{N} \sum_{m=r}^{r+d}\cos\left ( \varphi (x,y,\theta_{i},m) \right ),
\label{eq:ARF}
\end{equation}
where $N$ is the number of support region lines as described previously, $d$ corresponds to the width of the ring (band), and $R_{max}$ represents the outer limit of the band.

The result of applying equation~\ref{eq:ARF} on the input image $I$ is the filter's response image.  For each candidate point we can obtain the radius of support region at that location. The shape estimation is performed by searching for the radius of the ring support region for each candidate in the image:
\begin{equation}
R_{ARF}(x,y)= \underset{0\leq r\leq R_{max}}{\operatorname{argmax}} \frac{1}{N\times d} \sum_{i=1}^{N} \sum_{m=r}^{r+d}\cos\left ( \varphi (x,y,\theta_{i},m) \right ),
\label{eq:ARFShape}
\end{equation}
where $R_{ARF}$ is the radius of the support region that corresponds to the highest convergence for location $(x,y)$.

The SBF support region shown in \figurename{~\ref{fig:SBF}} is a band of fixed width with varying radius in each direction, where the maximization of the convergence index at each point is obtained by: 
\begin{equation}
F_{SBF}(x,y) = \frac{1}{N}\sum_{i=1}^{N}\left [ \max\limits_{R_{min}\leq r\leq R_{max}}\sum_{m=r}^{r+d}\cos\left ( \varphi (x,y,\theta_{i},m) \right ) \right ],
\label{eq:SBF}
\end{equation}
where $R_{min}$ and $R_{max}$ represent the inner and outer sliding band limits, respectively. 

Given the more flexible shape formulation for the SBF support region, the shape is defined by $N$ independent radii:
\begin{equation}
\begin{split}
&R_{SBF}(x,y,i) =  \underset{R_{min}\leq r\leq R_{max}}{\operatorname{argmax}}\left [\sum_{m=r}^{r+d}\cos\left ( \varphi (x,y,\theta_{i},m) \right ) \right ],
\label{eq:SBFShape}
\\
&\overline{R_{SBF}}(x,y) = \frac{1}{N}\sum_{i=1}^{N}R_{SBF}(x,y,i),
\end{split}
\end{equation}	
where $R_{SBF}(x,y,i)$ has a different value for each direction and $\overline{R_{SBF}}(x,y)$ represents the radius average.


\begin{figure*}[!t]
	\centering
	\captionsetup[subfigure]{labelformat=empty}
	\subfloat{\includegraphics[trim={1.30cm 1.2cm 1.30cm 1.0cm},clip,width=0.13\textwidth]{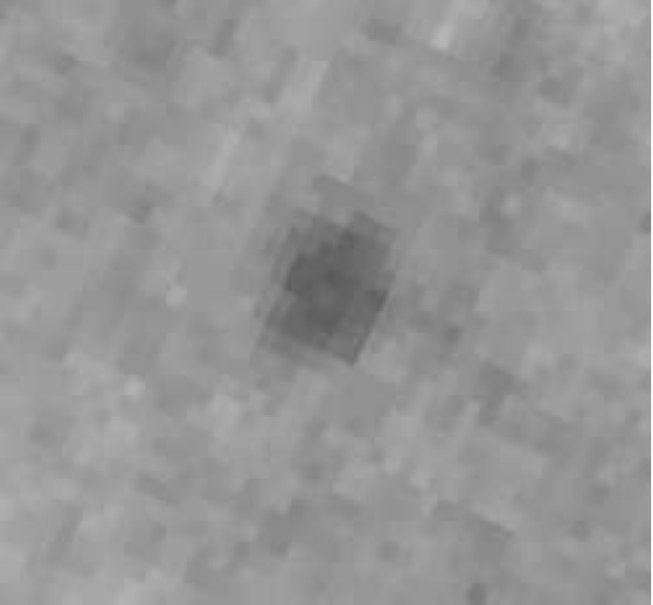}{\footnotesize \bluetext{ \put(-12,3){$I_{N}$}}}
		\label{fig:LCF-Green}}
	\hfill
	\subfloat{\includegraphics[trim={1.30cm 1.1cm 1.30cm 1.1cm},clip,width=0.13\textwidth]{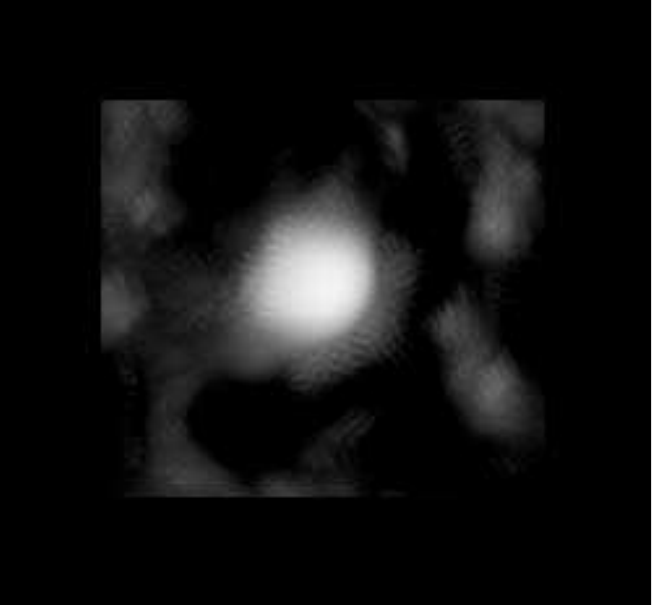}
		\label{fig:LCF-GW-ARF-Norm}}
	\hfill
	\subfloat{\includegraphics[trim={1.30cm 1.1cm 1.30cm 1.1cm},clip,width=0.13\textwidth]{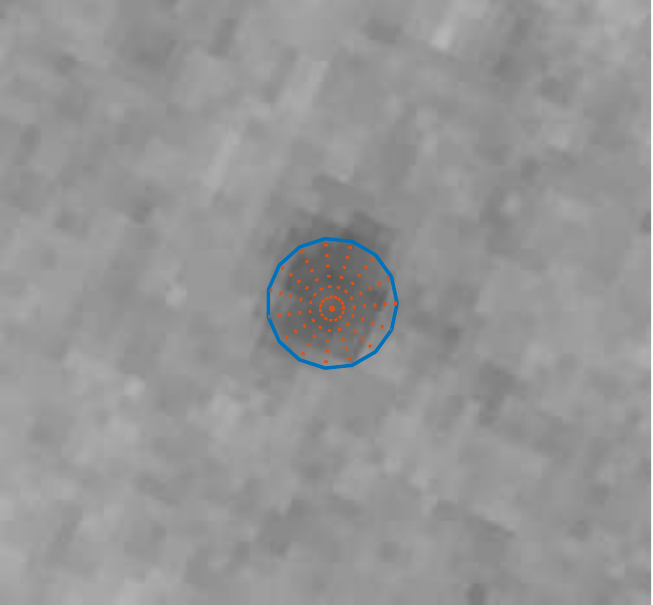}
		\label{fig:LCF-GW-R-ARF-Norm}}
	\hfill
	\subfloat{\includegraphics[trim={1.30cm 1.1cm 1.30cm 1.1cm},clip,width=0.13\textwidth]{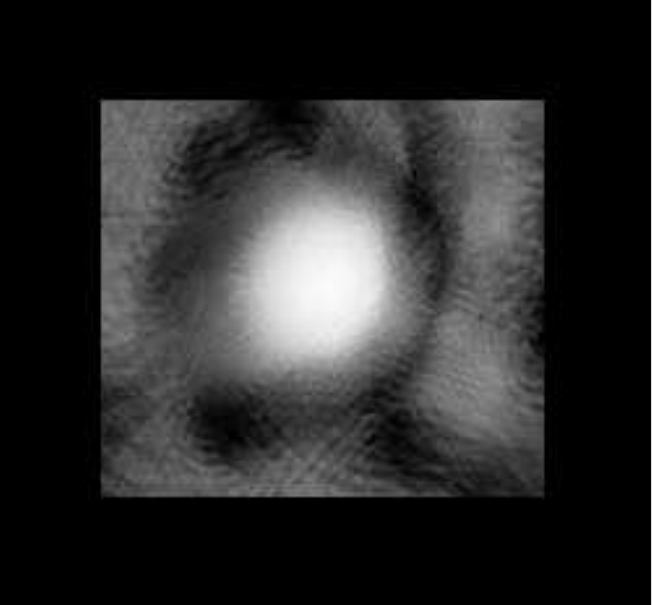}
		\label{fig:LCF-GW-SBF-Norm}}
	\hfill
	\subfloat{\includegraphics[trim={1.30cm 1.1cm 1.30cm 1.1cm},clip,width=0.13\textwidth]{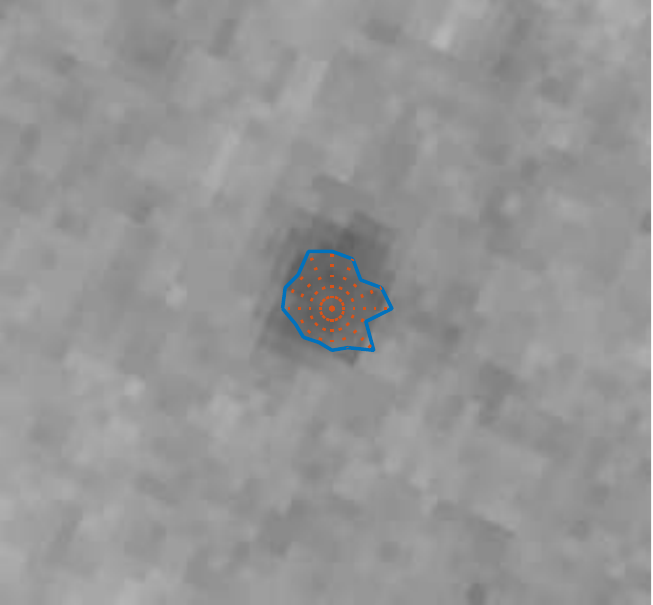}
		\label{fig:LCF-GW-R-SBF-Norm}}
	\hfill
	\subfloat{\includegraphics[trim={1.30cm 1.1cm 1.30cm 1.1cm},clip,width=0.13\textwidth]{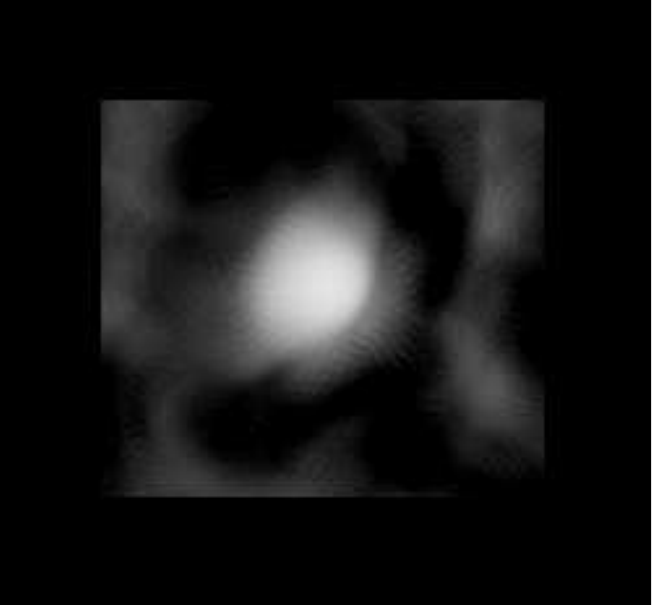}
		\label{fig:LCF-GW-SEF-Norm}}
	\hfill
	\subfloat{\includegraphics[trim={1.30cm 1.1cm 1.30cm 1.1cm},clip,width=0.13\textwidth]{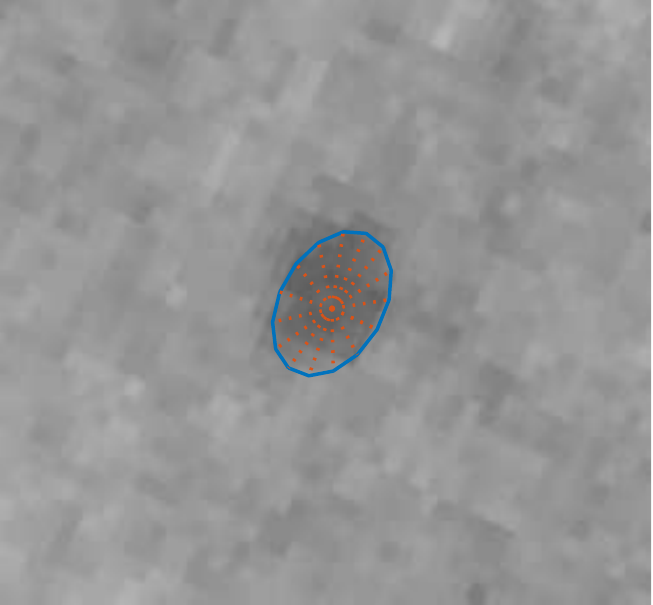}
		\label{fig:LCF-GW-R-SEF-Norm}}
	
	\subfloat[(a) input patch]{\includegraphics[trim={1.30cm 1.1cm 1.30cm 1.1cm},clip,width=0.13\textwidth]{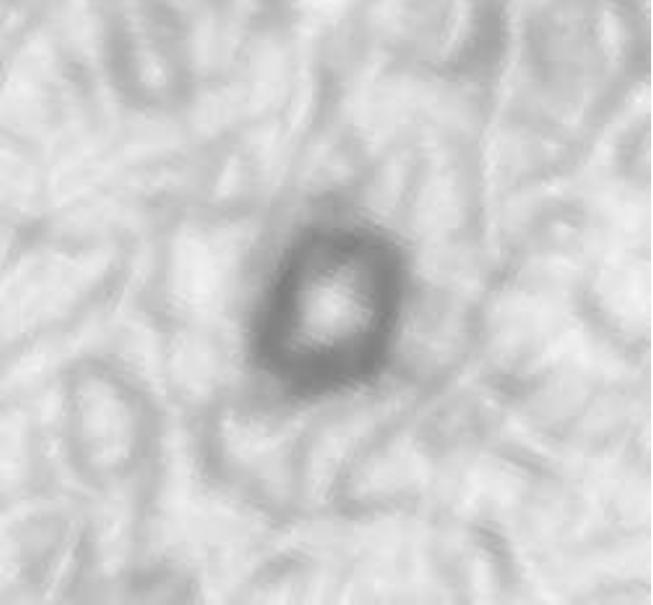}{\footnotesize \bluetext{ \put(-24,3){$I_{WOS}$}}}
		\label{fig:LCF-GW}}
	\hfill
	\subfloat[(b) $F_{ARF}$]{\includegraphics[trim={1.30cm 1.1cm 1.30cm 1.1cm},clip,width=0.13\textwidth]{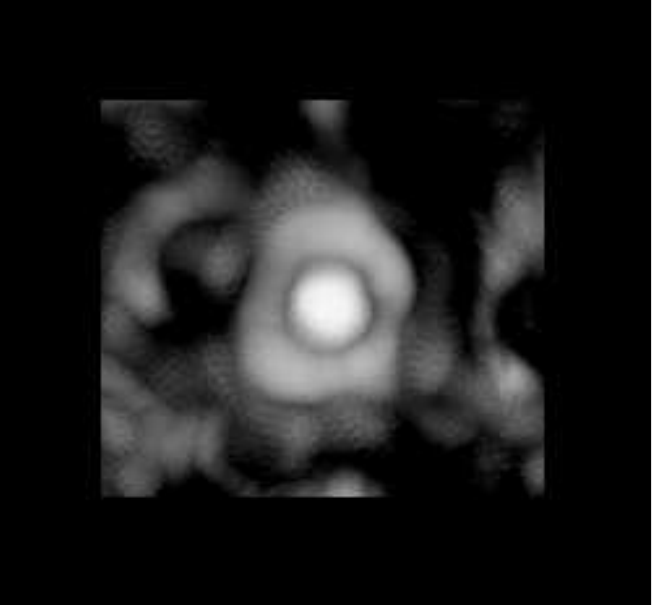}
		\label{fig:LCF-GW-ARF}}
	\hfill
	\subfloat[(c) $R_{ARF}$]{\includegraphics[trim={1.30cm 1.1cm 1.30cm 1.1cm},clip,width=0.13\textwidth]{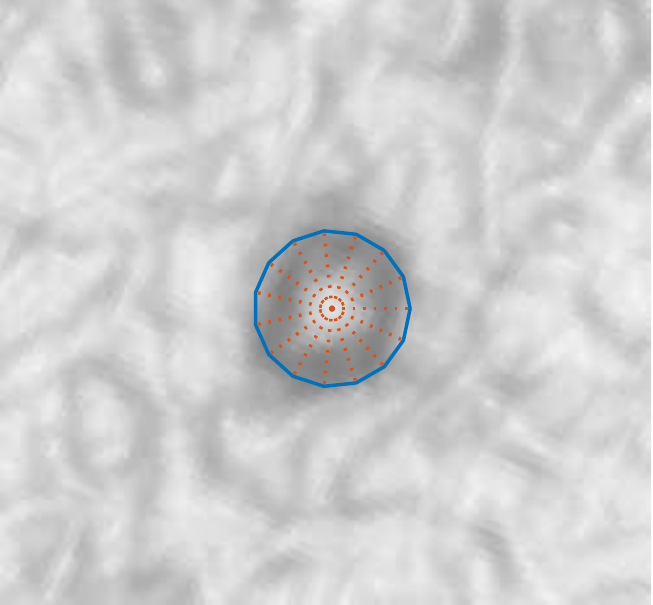}
		\label{fig:LCF-GW-R-ARF}}
	\hfill
	\subfloat[(d) $F_{SBF}$]{\includegraphics[trim={1.30cm 1.1cm 1.30cm 1.1cm},clip,width=0.13\textwidth]{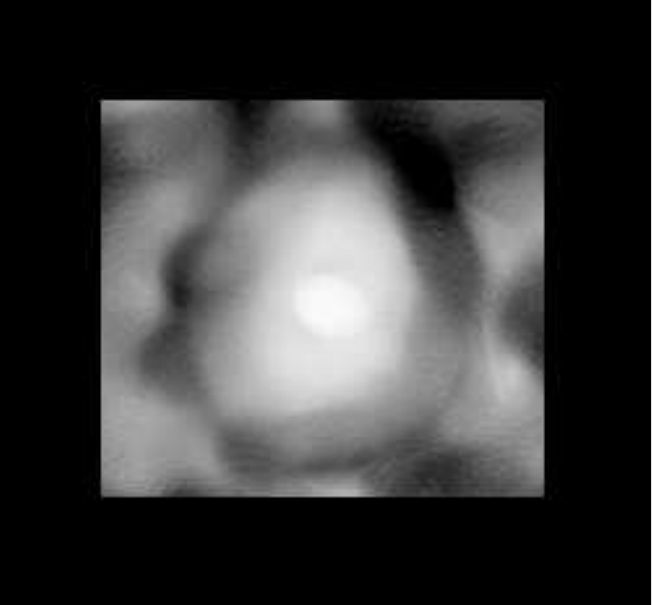}
		\label{fig:LCF-GW-SBF}}
	\hfill
	\subfloat[(e) $R_{SBF}$]{\includegraphics[trim={1.30cm 1.1cm 1.30cm 1.1cm},clip,width=0.13\textwidth]{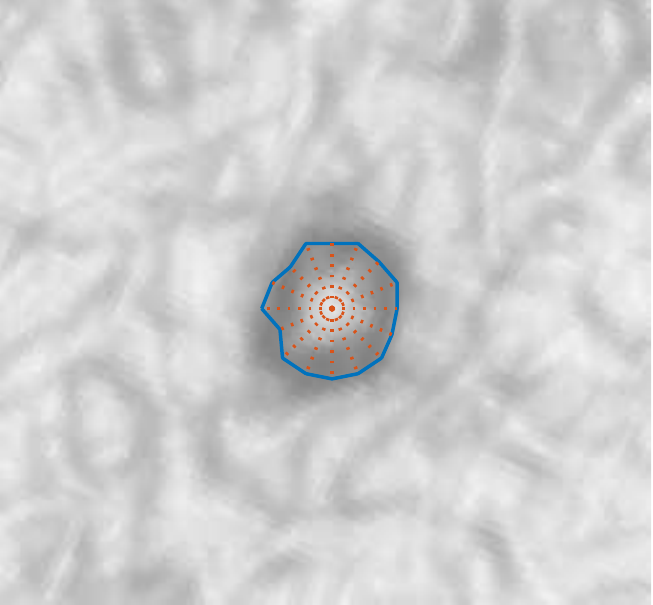}
		\label{fig:LCF-GW-R-SBF}}
	\hfill
	\subfloat[(f) $F_{SEF}$]{\includegraphics[trim={1.30cm 1.1cm 1.30cm 1.1cm},clip,width=0.13\textwidth]{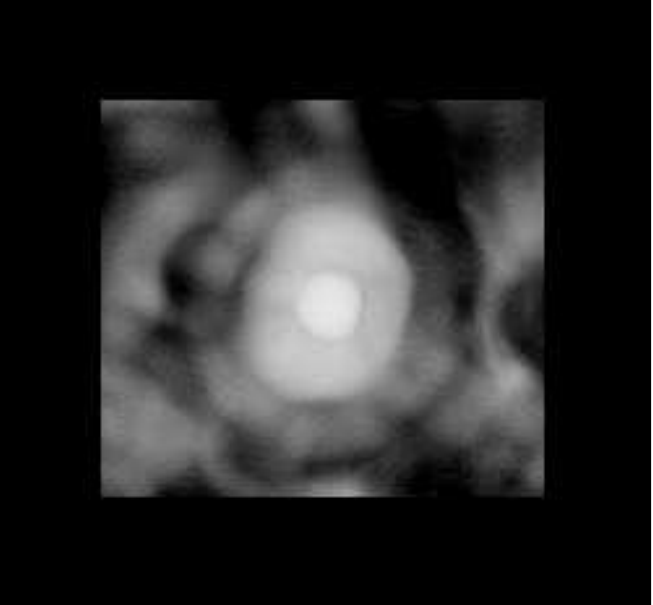}
		\label{fig:LCF-GW-SEF}}
	\hfill
	\subfloat[(g) $R_{SEF}$]{\includegraphics[trim={1.30cm 1.1cm 1.30cm 1.1cm},clip,width=0.13\textwidth]{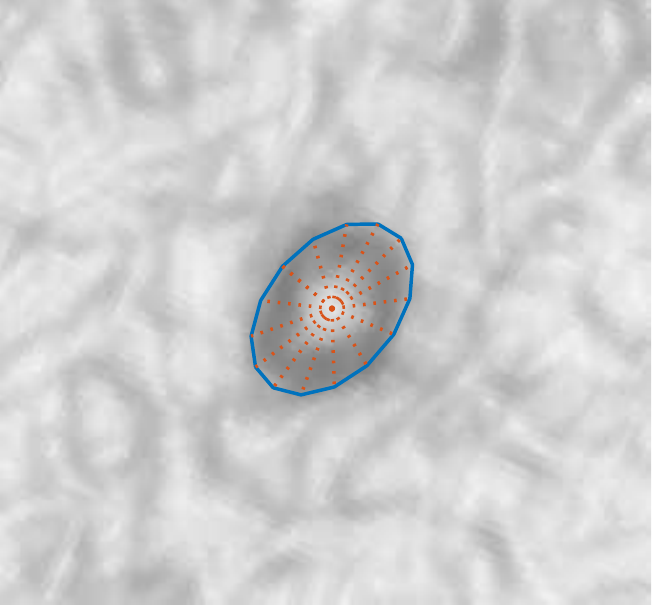}
		\label{fig:LCF-GW-R-SEFtt}}
	\vspace{-0.5em}
	\caption{The local convergence index filters' responses and  corresponding estimated shapes and radii on a normalized image patch ($I_N$) in the \nth{1} row and on a gradient-weighted image patch ($I_{WOS}$) in the \nth{2} row.  The patch size is $50 \times 50\;px$.}
	\label{fig:LCF-Sampel}
	\vspace{-1.5em}
\end{figure*}

The super-elliptical band in the SEF filter  allows the model to characterize a larger variety of shapes whilst at the same time reducing the irregularity in shape by its 2-fold symmetry constraints.
The SEF filter is defined by
\begin{equation}
{\small \begin{split}
	&F_{SEF}(x,y) = \frac{1}{N} \max\limits_{1\leq j\leq N/4}\Biggl[     \max\limits_{R_{min}\leq r\leq R_{max}} \frac{1}{2d} \sum_{m=r}^{r+d} \big ( \cos (\varphi_{(j,m)})
	\\&+\cos (\varphi_{(j+\frac{N}{2},m)}) \big )
	\\ &+ \max\limits_{R_{min}\leq r\leq R_{max}}\frac{1}{2d} \sum_{m=r}^{r+d}\left ( \cos (\varphi_{(j+\frac{N}{4},m)})+\cos (\varphi_{(j+\frac{3N}{4},m)}) \right )
	\\ &+  \sum_{i=1}^{N/4-1}\max\limits_{R_{min}\leq r\leq R_{max}}\frac{1}{4d} \sum_{m=r}^{r+d} \Big  ( \cos (\varphi_{(j+i,m)})+\cos (\varphi_{(j-i+\frac{N}{2},m)})
	\\&+ \cos (\varphi_{(j+i+\frac{N}{2},m)})+\cos (\varphi_{(j-i+N,m)}) \Big )\Biggr],
	\end{split}}
\label{eq:SEF1}
\end{equation}
To simplify the equation, we used  $\varphi_{(i,m)}$ instead of $\varphi(x,y,\theta_{i},m)$ which represents the angle between the gradient vector at point ($i,m$) and the direction that is currently being analyzed. 
In order to consider the possible orientations of the super-elliptical filter, the parameter $j$ is introduced in this filter as shown in \figurename{~\ref{fig:SEF}}.
The shape estimation in the SEF is more complex than in the case of the ARF and SBF filters. Here, the $R_{SEF}(x,y)$ is calculated by averaging the major and minor radii of the obtained super-ellipse.

The described  ARF, SBF and SEF filters are applied on the normalized image ($I_N$) separately, and the filters' responses and estimated radii at the center of each candidate are included in the set of features. 
The same process is repeated by applying the filters on the gradient-weighted image ($I_{WOS}$). In total 12 LCF-based features are extracted as described in Table~\ref{tab:featuresList}. The local convergence index filters' responses and corresponding estimated shapes and radii on a normalized image patch ($I_N$) are shown in the first row of \figurename{~\ref{fig:LCF-Sampel}} while the second row illustrates the results on the gradient-weighted image ($I_{WOS}$).

The complete feature set $\mathbb{F}$ for the final classification step includes 29 features, which are 7 intensity-based features, 10 shape-based descriptors and 12 LCF-based features as shown in Table~\ref{tab:featuresList}. For each candidate in the $\mathbb{M}$ set, all the features are extracted and  the entire feature set is rescaled to a  normal distribution with zero mean and unit standard deviation.

\subsection{Supervised Classification}\label{sec:SupervisedClassification}

To discriminate MAs from non-MA candidates, we use a hybrid sampling/boosting algorithm, called RUSBoost, proposed by Seiffert \textit{et al.}~\cite{Seiffert2010RUSBoost}.
RUSBoost is an adaptive boosting classifier (AdaBoost)~\cite{ freund1996experiments} in combination with a random undersampling technique (RUS)~\cite{Van2007Experimental} which is specifically designed to improve the performance of models trained on imbalanced data.

Boosting algorithms combine weak classifiers into a single ensemble classifier through majority voting. In particular, the AdaBoost algorithm~\cite{ freund1996experiments} starts by setting all instances' weights equally. Then, in each iteration, a weak model is formed by the base learner and then based on the calculated error the weights are adjusted in a manner that the weights of misclassified instances are increased while the weights of correctly classified instances are decreased. Weak models are added sequentially in each iteration and trained using the weighted data to correctly classify previously misclassified instances. The process continues until a defined number of weak learners have been created or no further improvement can be made on the training set.

Random undersampling (RUS) is a technique to deal with the imbalanced class problem by changing the class distribution of the training data set~\cite{Van2007Experimental}. The sampling is performed simply by randomly discarding instances from the majority class until a specific class distribution is obtained. 
In particular, the RUSBoost randomly undersamples a subset from the majority class in each iteration of the AdaBoost algorithm and consequently, the weak learners are trained using a balanced training set. 
For new input data, each weak learner generates a prediction value, which is weighted by the learners’ stage value. The final class is assigned by the summation of all weighted prediction values. 

Within the context of MAs classification, RUSBoost (with decision trees as the weak learners) is a suitable classifier since we deal with a skewed set with the minority of MA candidates and the majority of non-MA candidates.

The details of RUSBoost classifier are given in  Algorithm~\ref{alg:RUSBoost}. 
The algorithm takes as input a training set $(F_1,L_1),...,(F_m,L_m)$, where  $F_i \in \mathbb F$ is a feature vector and $l_i \in L$ is its label.
The weight for each sample $i$ in iteration $t$ is denoted by $\mathcal{W}_t(i)$, where initially all weights are set to $1/m$. In each iteration, a temporary class-balanced training set $\mathbb{D}'$ is obtained by randomly undersampling the $\mathbb{D}$ set. 
At each round, the weak learner algorithm searches for $h_t$ that minimizes the error with respect to the selected subset $\mathbb{D}'$ and distribution weights $\mathcal{W}'$.
Then the weights  are updated by calculating the overall error $\epsilon_{t}$ and the weight update parameter $\alpha_t$.
The new weight distribution $\mathcal{W}_{t+1}$ is used to train the next weak learner, and the process is iterated $T$ times.
The final strong classifier $H(x)$ is a weighted combination of $T$ weak classifiers which gives us the probability of being MA for candidate $x$.

\begin{algorithm}[!t]
	\caption{RUSBoost classifier}
	\label{alg:RUSBoost}
	{\footnotesize 	
		\begin{algorithmic}
			\STATE \textbf{Input:} Set $\mathbb{D}$ of training samples $({F}_{1},l_{1}),...,({F}_{m},l_{m})$  where $m$ is the number of training samples, ${F}_{i}\in\mathbb{F}$ and  $l_{i}\in L=\{-1,1\}$
			\STATE \textbf{Output:} RUSBoost classifier $H(x)$
			\STATE \textbf{Initialize:} $\mathcal{W}_{1}(i) = \frac{1}{m}$ for all $i$
			\FOR{$t = 1$ to $T$ \textbf{step} 1} 
			
				\STATE$\mathbb{D}'\leftarrow \textrm{random undersampling (RUS) training set } \mathbb{D}$ 
				\STATE$\mathcal{W}_{t}'\leftarrow \textrm{extract weights for the subset }  \mathbb{D}'$
				\STATE Call \textit{WeakLearn} with subset $\mathbb{D}'$ and weights $\mathcal{W}_{t}'$  and get back weak classifier $h_{t}$:
					\STATE $\;\;\;\;\;\;\;h_t =\underset{h_j\in\mathbb{H}}{\operatorname{argmin}}\sum_{1}^{|\mathcal{W}_t'|} \mathcal{W}_t'(i)[l_i \neq h_j(F_i)]$ 
				
				\STATE Psedo-loss calculation for $\mathbb{D}$ and $\mathcal{W}_t$:
					\STATE  $\;\;\;\;\;\;\; \epsilon_{t} = \sum_{i=1}^{m} \mathcal{W}_t(i) [l_i \neq h_t(F_i)]$
				
				\STATE Weight update parameter:
					\STATE $\;\;\;\;\;\;\; \alpha_t= \frac{1}{2}\ln\frac{1-\epsilon_t}{\epsilon_{t}}$
				
				\STATE Update weights and normalization:
					\STATE $\;\;\;\;\;\;\;\mathcal{W}_{t+1}(i)={\mathcal{W}_t(i)\exp(-\alpha_t l_i h_t(F_i))}/{\sum_{i=1}^{m}\mathcal{W}_{t+1}(i)}$
			\ENDFOR				
				\STATE Output and final classifier:
					\STATE $\;\;\;\;\;\;\;H(x) = {\sum_{t=1}^{T}\alpha_{t} h_{t}(x)}/{\sum_{j=1}^{T}\alpha_{t}} $		
		\end{algorithmic}
	}
\end{algorithm}

\begin{figure}[!t]
	\centering
	\captionsetup[subfigure]{labelformat=empty}

	\subfloat[]{\includegraphics[trim={2.5cm 2cm 0.5cm 1.0cm},clip,width=2.10in]{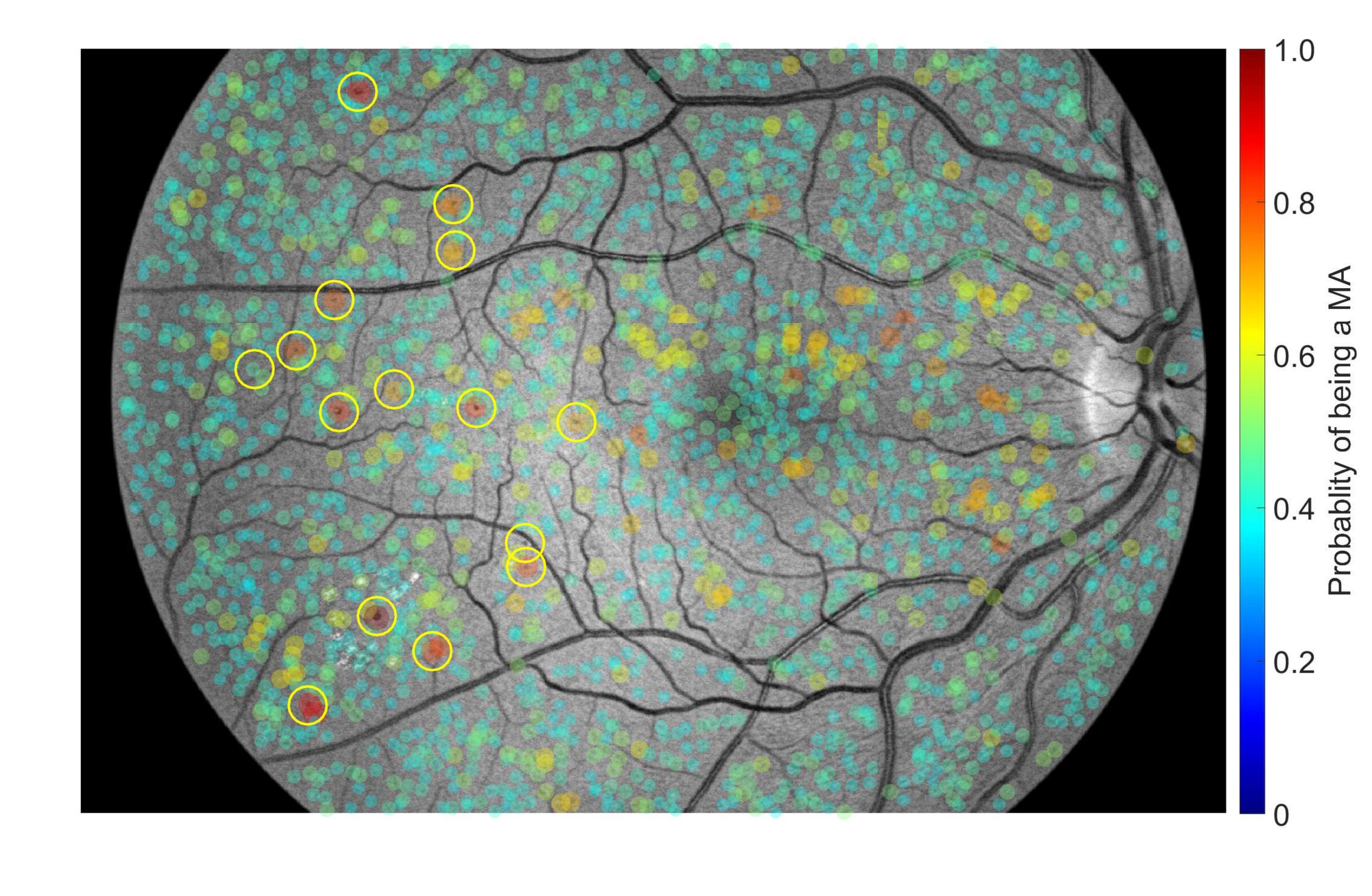}%
	\label{fig:Classification:wholeimage2}}
\hfil
\subfloat[]{\includegraphics[trim={0cm 1.2cm 3cm 1.0cm},clip,width=1.3in]{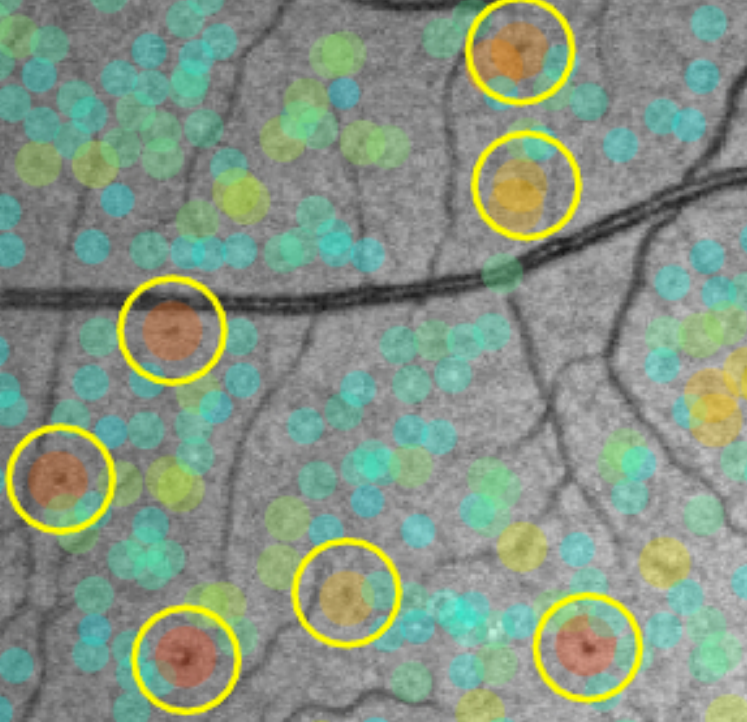}%
	\label{fig:Classification:cropimage2}}%
	\vspace{-1.4em}
	\caption{Example of classification result in which the probabilities of being MA  for each candidate are shown using a heat color map. The yellow circles represent the annotation by experts.}
	\label{fig:classification}
	\vspace{-1.5em}
\end{figure}

\figurename{~\ref{fig:classification}} shows examples of a RUSBoost classifier result in which the probability of being MA for each candidate is indicated using a heat color map. The yellow circles represent the annotation by experts.


\section{Validation and Experimental Results}\label{sec:Validation}
\subsection{Material}\label{sec:Material}
For the evaluation of proposed method, we used three publicly available datasets, i.e. the  e-ophtha-MA~\cite{e-ophtha-MA}, DiaretDB1~\cite{diaretdb1} and Retinopathy Online  Challenge training set (ROC)~\cite{niemeijer2010retinopathy} and also our two new public RetinaCheck datasets called RC-RGB-MA and RC-SLO-MA~\cite{RC-RGB-MA,RC-SLO-MA}.
The specifications of these datasets are summarized in Table~\ref{tab:Datasets} including the imaging modality, image resolution, camera angle view and number of annotated MAs.

 \begin{table*}[!t]
 	\renewcommand{\arraystretch}{0.9}
 	\centering
 	\caption{Datasets specifications.}
 	\vspace{-0.5em}
 	{\footnotesize 	
 		\begin{tabular}{lC{4em}C{12em}C{3.5em}C{6em}C{4.5em}C{4.5em}C{1.7em}C{1.7em}C{1.7em}}
 			\toprule
 			
 			Dataset	& Modality & Image size ($px$) & FOV (degree) & FOV diameter ($px$) & Pixel size ($\mu m/px$)   & Number of experts & $N_{IN}$ & $N_{IP}$ & $N_{MA}$\\	
 			
 			\midrule
 			\midrule
 			e-ophtha-MA~\cite{e-ophtha-MA} & RGB & $1440 \times 960$ - $2544 \times 1696$ & $45^{\circ}$ & 910 - 1925 & $\sim$ 7-15  & 1 & 233 & 148 & 1306\\
 			RC-RGB-MA~\cite{RC-RGB-MA} & RGB & $2592 \times 1944$ & $45^{\circ}$ & 2087 & $\sim 7$  & 2 &	81$^\dagger$ & 99$^\dagger$ & 342$^\dagger$\\
 			RC-SLO-MA~\cite{RC-SLO-MA} & SLO & $1024 \times 1024$ & $45^{\circ}$ & 1024 & $\sim 14$   &2&  10$^\dagger$ & 44$^\dagger$ & 114$^\dagger$\\
 			DiaretDB1~\cite{diaretdb1} & RGB & $1500 \times 1152$ & $50^{\circ}$& 1415 & $\sim 11$ &4  & 50$^\ddagger$ & 39$^\ddagger$ & 182$^\ddagger$\\
 			ROC~\cite{niemeijer2010retinopathy} & RGB & $768 \times 576$ - $1394 \times 1392$ & $45^{\circ}$ & 720 - 1345 & $\sim$ 11-20& 4  & 13 & 37 & 336\\
 			\bottomrule
 			\noalign{\vskip 1mm}
 			\multicolumn{10}{l}{$N_{IN}$: number of images without MAs; $N_{IP}$: number of images with MAs; $N_{MA}$: total number of MAS.}\\
 			\multicolumn{10}{l}{$^\dagger$based on two experts agreement; $^\ddagger$ confidence level higher than 0.75.}
 		\end{tabular}
 	}
 	\label{tab:Datasets}%
 \end{table*}%
 
 \begin{table*}[!t]
 	\renewcommand{\arraystretch}{0.9}
 	\centering
 	\caption{Agreement between experts on the annotations of the RC-RGB-MA and RC-SLO-MA datasets and DiaretDB1 ground truth confidence levels.}
 	\vspace{-0.5em}
 	{\footnotesize 
 		\begin{tabular}{p{2.9cm}C{1.50em}C{1.50em}C{1.50em}cp{2.9cm}C{1.50em}C{1.50em}C{1.50em}cp{2cm}C{1.50em}C{1.50em}C{1.50em}}
 			\toprule
 			\multicolumn{4}{c}{RC-RGB-MA (250)}	&&	\multicolumn{4}{c}{RC-SLO-MA (58)}&&	\multicolumn{4}{c}{DiaretDB1 (89)}\\
 			\cline{1-4}\cline{6-9}\cline{11-14}
 			\noalign{\vskip 1mm}   
 			Expert& $N_{IN}$ & $N_{IP}$ & $N_{MA}$ 	&&Expert&  $N_{IN}$ & $N_{IP}$ & $N_{MA}$	&&$conf_{GT}$&  $N_{IN}$ & $N_{IP}$ & $N_{MA}$\\
 			\midrule
 			\midrule
 			\nth{1} expert  	& 132   & 118   & 537 	&&\nth{1} expert  (Green)& 11   & 47   & 213&&25\%       & 20    & 69    & 870 \\
 			\nth{2} expert   & 100   & 150   & 691 	&&\nth{2} expert (Infrared)& 12   & 46   & 178&&50\%        & 38    & 51    & 505 \\
 			Two-agreement  				& 81    & 99    & 342 		&&Two-agreement& 10    & 44    & 114&&75\%       & 50    & 39    & 182 \\
 			\nth{2} expert success rate & 61.3\% &83.9\% & 63.7\% 	&&\nth{2} expert success rate&  99.9\% & 93.6\% & 53.5\% &&100\%        & 76    & 13    & 37 \\
 			
 			\bottomrule
 			\noalign{\vskip 1mm}
 			\multicolumn{14}{l}{$N_{IN}$: number of images without MAs; $N_{IP}$: number of images with MAs; $N_{MA}$: total number of MAS; $conf_{GT}$: confidence level.}\\
 		\end{tabular}
 	}
 	\label{tab:DRSAgreement3datasets}%
 	\vspace{-1.5em}
 \end{table*}%

\subsubsection{e-ophtha-MA} 

e-ophtha is a public database of color fundus images designed for scientific research in Diabetic Retinopathy~\cite{e-ophtha-MA}. It contains 233 healthy images (i.e. no lesion) and 148 images with microaneurysms or small hemorrhages manually annotated by ophthalmologists. The images have four different resolutions, ranging from $1440 \times 960$ to $2544 \times 1696$ pixels with $45^{\circ}$ field of view (FOV).
\subsubsection{RC-RGB-MA}
The RC-RGB-MA is a retinal image dataset which is collected in the framework of the RetinaCheck project managed by  Eindhoven University of Technology, the Netherlands~\cite{RC-RGB-MA}. The 250 RGB images in this dataset are acquired with a DRS non-mydriatic fundus camera with a resolution of $2595 \times 1944$ and $45^\circ$ FOV. 
Two experts annotated the MAs in all images using the Microaneurysm Annotation Tool (RC-MAT)~\cite{RC-MAT}. The agreement values between two experts are shown in Table~\ref{tab:DRSAgreement3datasets}. 
This dataset is used to assess the potential of the proposed method as a standalone application in a large-scale DR screening setting i.e. RetinaCheck project.

\subsubsection{RC-SLO-MA}
In addition to the RGB fundus images, we also provide a public dataset called RC-SLO-MA in which the images are captured using the Scanning Laser Ophthalmoscopy (SLO) technique~\cite{RC-SLO-MA}.
The images in this dataset are acquired with an EasyScan camera (i-Optics Inc., the Netherlands) using both green and  infrared lasers. The RC-SLO-MA dataset includes 58 images with a resolution of $1024 \times 1024$ and a $45^\circ$ field of view. The MAs are once annotated by an expert using only the green images and another expert manually labeled the MAs using the infrared laser images.  
{As we can see in Table~\ref{tab:DRSAgreement3datasets}, the agreement value between two experts on the labeled MAs is less than the value obtained for the RC-RGB-MA dataset. On the other hand, the agreement between two experts on the discrimination between images with MAs and images without MAs is higher using the SLO images compared to the RGB ones.
The overall agreement results show that MA detection is a challenging task for the human experts as well.}     

\subsubsection{DiaretDB1}
The DiaretDB1 is a publicly available dataset comprises 89 color fundus images~\cite{diaretdb1}.   
Four medical experts marked the MAs independently and reported confidence levels $\{<50\%,\geqslant50\%,100\%\}$ which are representing the certainty of the decision that a marked finding is correct (Table~\ref{tab:DRSAgreement3datasets}). Because of the disagreement between the four experts’ annotations, a consensus of agreement higher than 75\% is used to assign an MA label to a region (resulting in 182 MAs). There are 3 to 5 dark spot shaped artefacts caused by a dirty camera lens which are located in the exact same position in the images of the DiaretDB1 dataset. 

\subsubsection{ROC} 
The Retinopathy Online Challenge (ROC)~\cite{niemeijer2010retinopathy} contains 50 training images and 50 test images, where the annotations for the training set are publicly available. Since for the test images the gold standard is not provided and a new submission to this challenge is no longer possible, we use only the 50 training images to train and test with a 10-fold cross validation approach.

\subsection{Parameters Optimization}
As we described in Section~\ref{sec:SupervisedClassification}, in the final classification step we use the RUSBoost algorithm with the decision trees as its weak learners. The implementation requires the determination of a number of parameters including:

\begin{enumerate}
	\item {Maximum number of splits:} a parameter required for the construction of the decision trees. For a maximum number of splits equal to 1, each node represents a stump. By increasing the number of splits, the complexity of the decision trees will increase.
	\item {Learning rate $\lambda$:} a shrinking parameter which reduces the contribution of each weak learner in order to prevent overfitting. The optimal choice of the learning rate depends critically on the number of trees; a small $\lambda$ requires a large number of trees to achieve good performance~\cite{James2013}.
	\item {Number of trees:} is the number of trees used for training. Increasing the number of trees results in a decrease in both the bias and variance terms in the bias-variance decomposition. On the other hand, using too many trees may result in overfitting~\cite{James2013}.
\end{enumerate}

\begin{figure}[!t]
	\centering
	\includegraphics[trim={2.2cm 0cm 2.0cm 0.7cm},clip,width=0.49\textwidth]{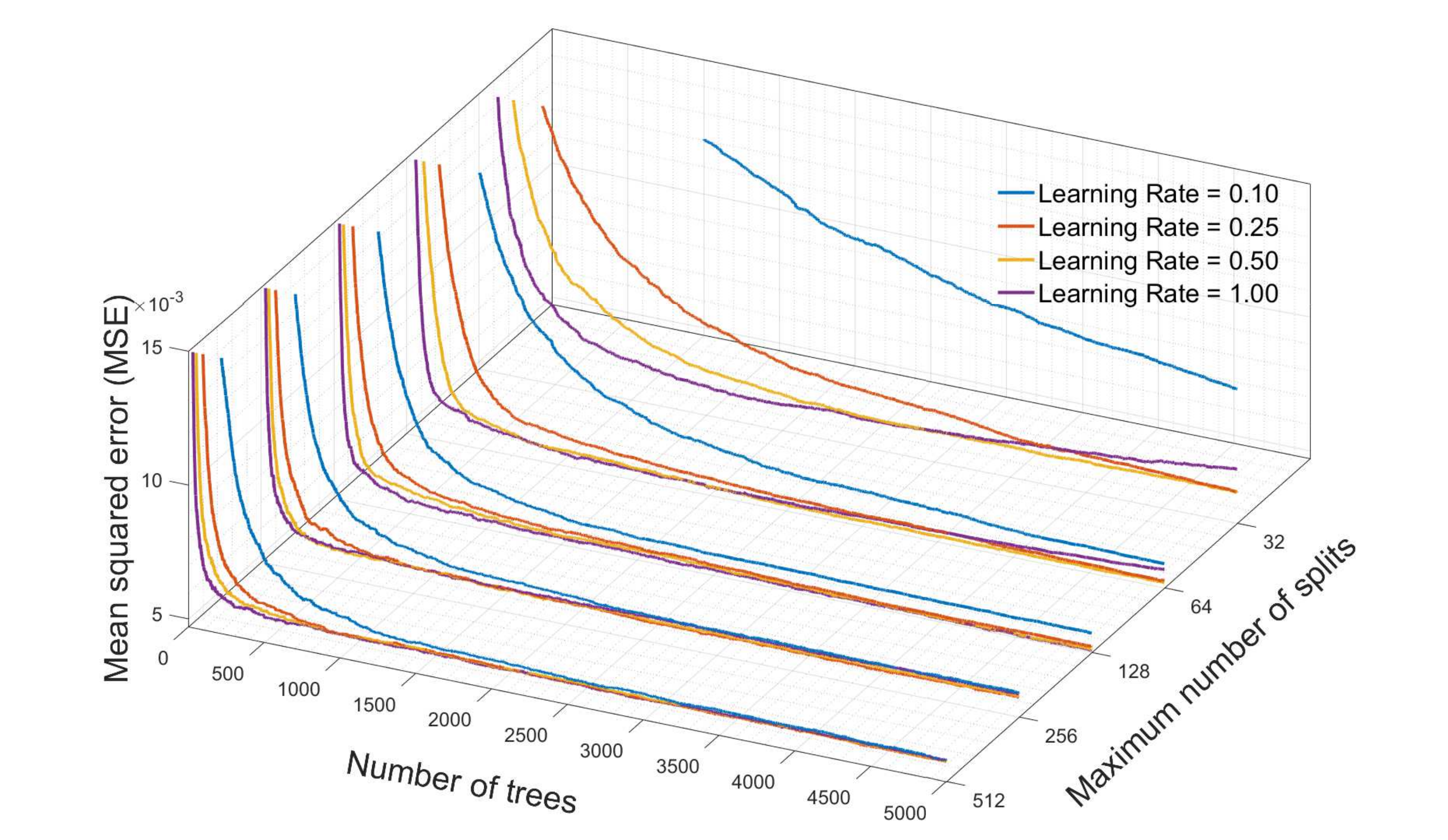}%
	\caption{RUSBoost parameters optimization by plotting the mean squared error curves for up to 5000 trees with the maximum number of splits from $32$ to $512$ and different learning rates ($\lambda = {0.1, 0.25, 0.5, 1}$).}
	\label{fig:ParametersOptimization}
	\vspace{-1.5em}
\end{figure}

To investigate which parameters would be optimal, we trained classification ensembles for up to 5000 trees, varied the maximum number of splits from 1 to 512 and used as learning rates ${0.1, 0.25, 0.5, 1}$. The optimization is done on 10\% of e-ophtha-MA samples using all extracted features. A part of this experiment is shown in \figurename{~\ref{fig:ParametersOptimization}}, demonstrating the mean squared error (MSE) for the maximum number of splits equal to ${32, 64, 128, 256, 512}$.

The optimal parameters obtained in this experiment were $4727$ for the number of trees, $128$ as the maximum number of splits and a learning rate of $0.5$. It is clear from \figurename{~\ref{fig:ParametersOptimization}} that for a large enough number of splits, the RUSBoost classifier does not seem to lead to overfitting, even in case $\lambda=1.00$. 
Using too many weak learners and a high number of splits leads to a long computation time but a small decrease in the MSE.
The curves in \figurename{~\ref{fig:ParametersOptimization}} show a very small decrease in the MSE values by increasing the number of trees higher than $1000$. Therefore, we use  $1000$ as the number of trees,  $100$ as the maximum number of splits and $0.5$ as the learning rate.


\subsection{Candidate Extraction Evaluation}

Table~\ref{tab:CandidateEvaluation} compares the sensitivity of the proposed MA candidate extraction technique with the previously published candidate extractor algorithms \cite{Lazar2013,Dai2016,Shah2016,Adal20141,Walter2007555,Zhang20102237,Abdelazeem2002} on the ROC training set. As demonstrated in Table~\ref{tab:CandidateEvaluation}, the proposed method achieves a sensitivity value of 0.82 which is higher than the other methods. Although the average of false positives per image (FPI) in the proposed method is higher than the others, the number of extracted candidates is still significantly smaller than the total number of pixels in the image ($\approx0.06\%$).  In the candidate extraction phase, the sensitivity values of 0.95, 0.94, 0.72 are 0.75 are obtained for the e-ophtha-MA, RC-RGB-MA, RC-SLO-MA and DiaretDB1 datasets, respectively.

%

\begin{table}[!t]
	\renewcommand{\arraystretch}{0.9}
	\centering
	\caption{Candidate extraction performance using the ROC dataset.}
	\vspace{-0.5em}
	{\footnotesize 	\begin{tabular}{p{3cm}C{2.5cm}C{2cm}}
			\toprule
			Method & Sensitivity & FPI  \\
			\midrule\midrule
			Proposed method& 			\textbf{0.82}&		755.50\\
			Dai \textit{et al.}~\cite{Dai2016} 			&			0.69&		569.39\\
			Lazar\textit{ et al.}~\cite{Lazar2013} 		&			0.60&		569.39\\
			Shah \textit{et al.}~\cite{Shah2016} 		&			0.48&		65.00\\
			Lazar \textit{et al.}~\cite{Lazar2013}  	&			0.48&		73.94\\		
			Adal \textit{et al.}~\cite{Adal20141}	 	&			0.45&		\textbf{35.20}\\	
			Walter \textit{et al.}~\cite{Walter2007555}	&			0.36&		154.42\\	
			Zhang \textit{et al.}~\cite{Zhang20102237} 	&			0.33&		328.30\\	
			Abdelazeem~\cite{Abdelazeem2002} 			&			0.28&		505.85\\
			
			\bottomrule
	\end{tabular}}
	\label{tab:CandidateEvaluation}%
	\vspace{-1.5em}
\end{table}%

\subsection{Microaneurysm Detection Evaluation}
\begin{figure*}[h]
	\captionsetup[subfigure]{labelformat=empty}
	\centering
	
	\subfloat[]{\includegraphics[trim={18.8cm 11cm 5cm 2.8cm},clip,height=0.165\textwidth]{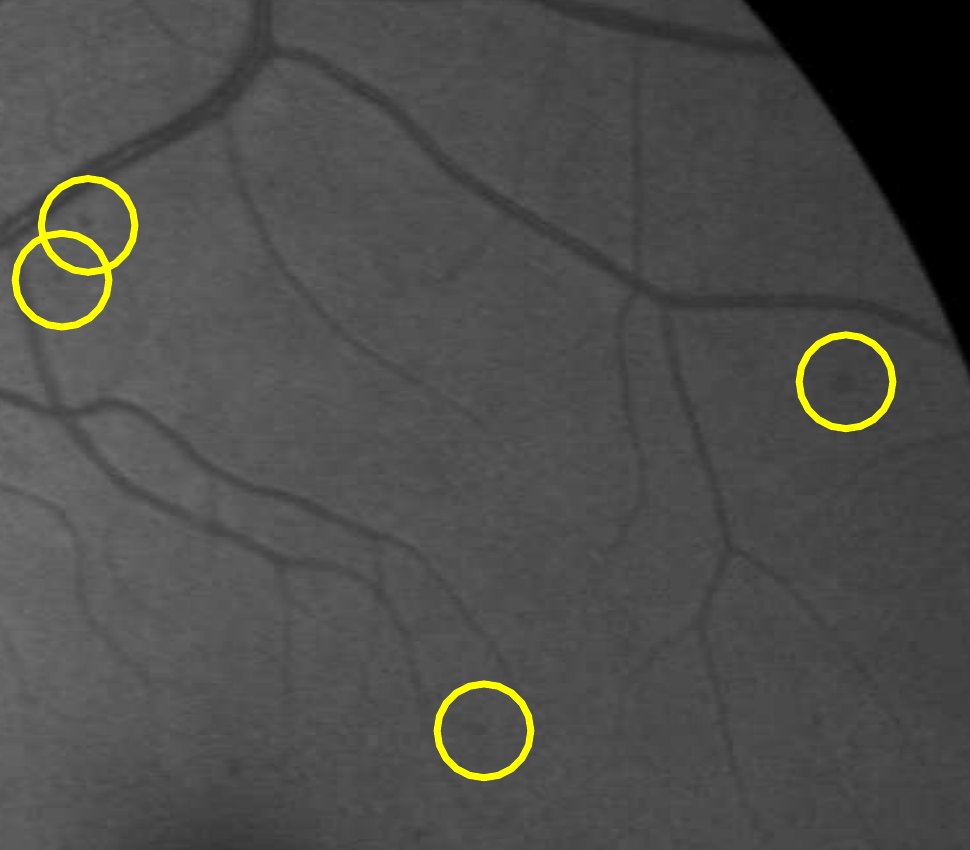}%
		\label{fig:eophthaSampleImageGT}}
	\subfloat[]{\includegraphics[trim={18.5cm 2cm 9cm 17.5cm},clip,height=0.165\textwidth]{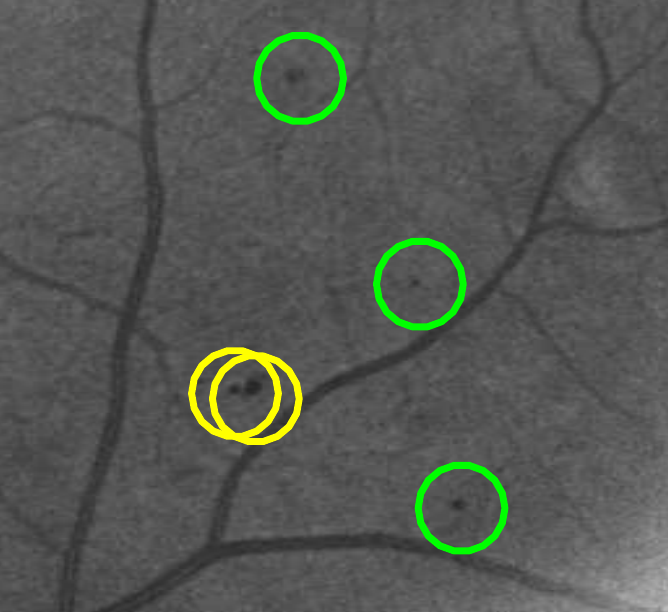}%
		\label{fig:DRSSampleImageGT}}
	\subfloat[]{\includegraphics[trim={11.0cm 7.0cm 1cm 5.0cm},clip,height=0.165\textwidth]{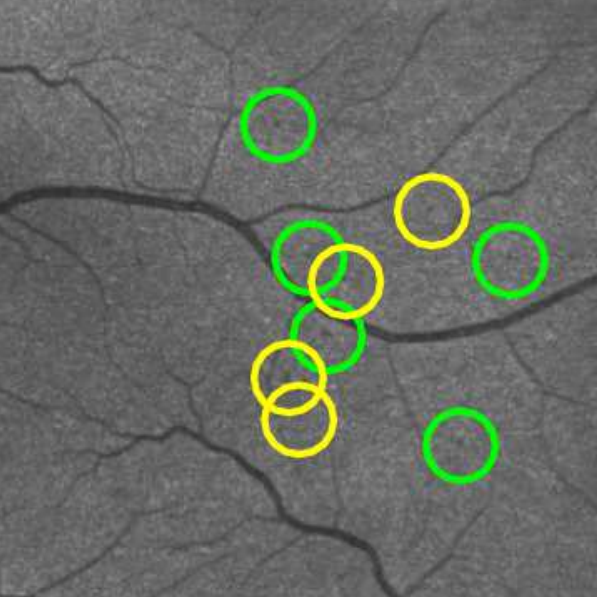}%
		\label{fig:SLOSampleImageGT}}
	\subfloat[]{\includegraphics[trim={6.0cm 4.0cm 12cm 10.0cm},clip,height=0.165\textwidth]{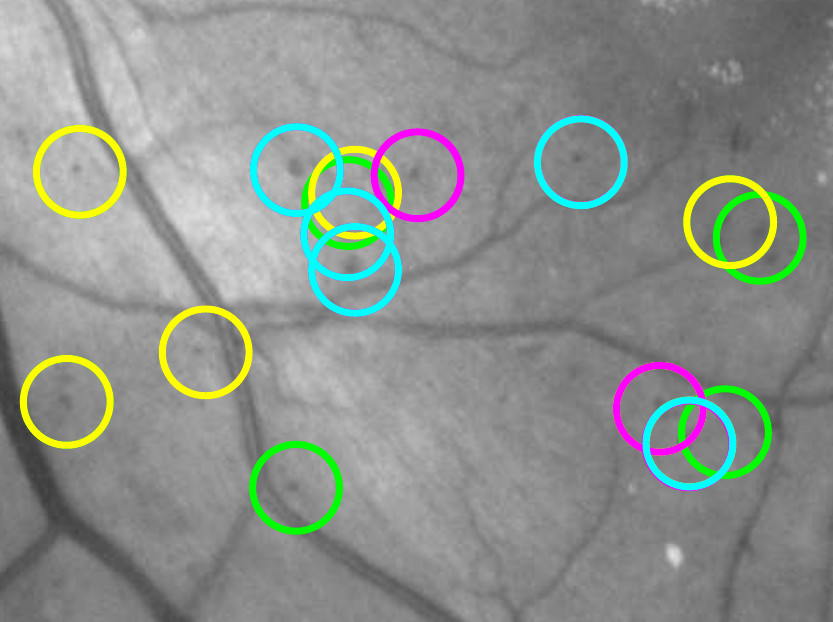}%
		\label{fig:Diretdb1SampleImageGT}}
	\subfloat[]{\includegraphics[trim={6.0cm 1.0cm 6cm 8.0cm},clip,height=0.165\textwidth]{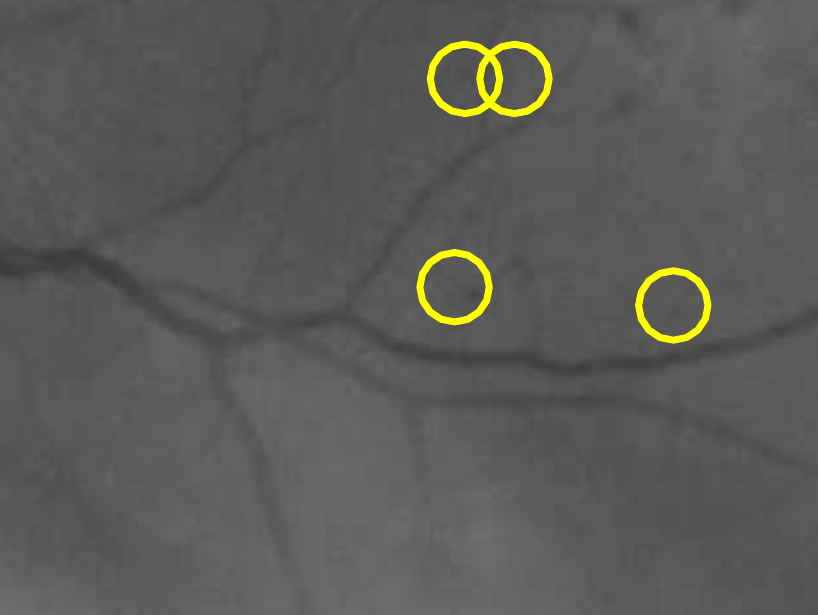}%
		\label{fig:ROCSampleImageGT}}
	\subfloat[]{\includegraphics[trim={0cm 0.20cm 0cm 0cm},clip,height=0.17\textwidth]{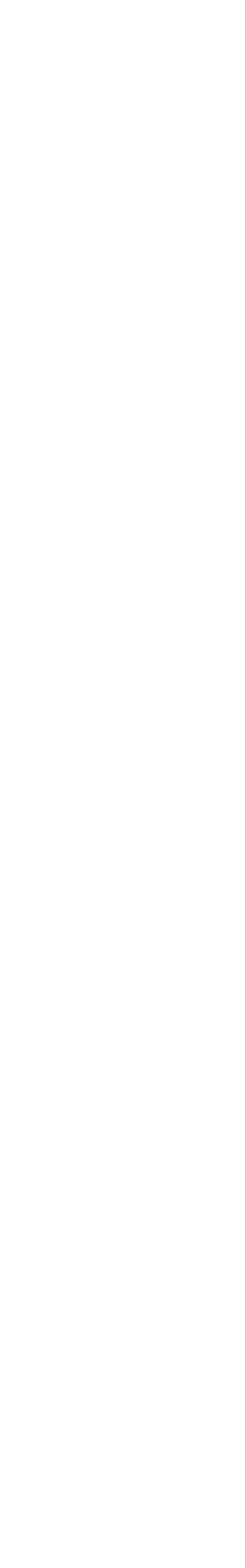}%
		\label{fig:blank}}
	\vspace{-2em}
	
	\subfloat[(a) e-Ophtha-MA]{\includegraphics[trim={18.8cm 11cm 5cm 2.8cm},clip,height=0.165\textwidth]{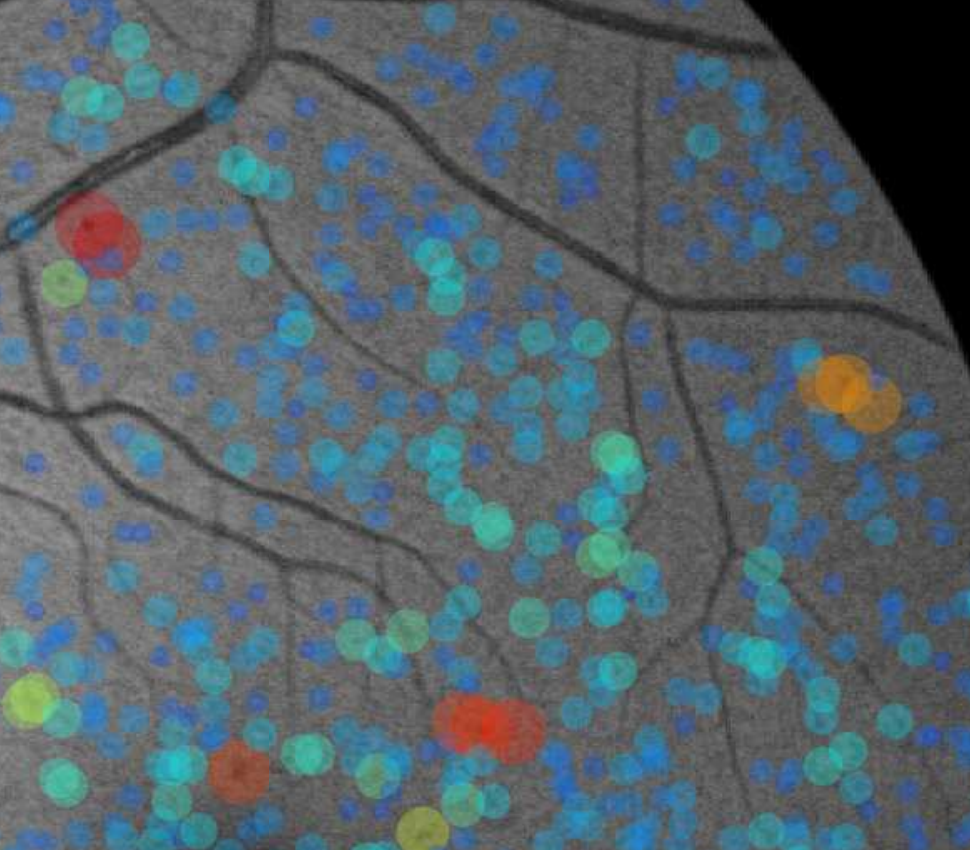}%
		\label{fig:eophthaSampleImageResult}}
	\subfloat[(b) RC-RGB-MA]{\includegraphics[trim={17.15cm 1.8cm 8.25cm 16.2cm},clip,height=0.165\textwidth]{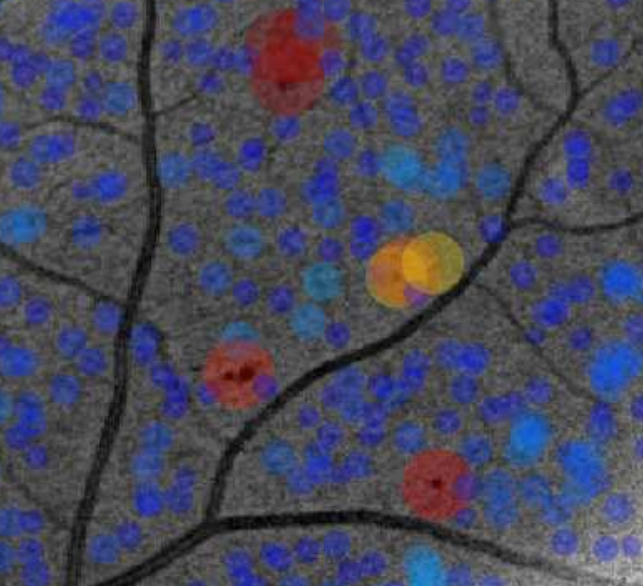}%
		\label{fig:DRSSampleImageResult}}
	\subfloat[(c) RC-SLO-MA]{\includegraphics[trim={12.0cm 7.5cm 1cm 5.5cm},clip,height=0.165\textwidth]{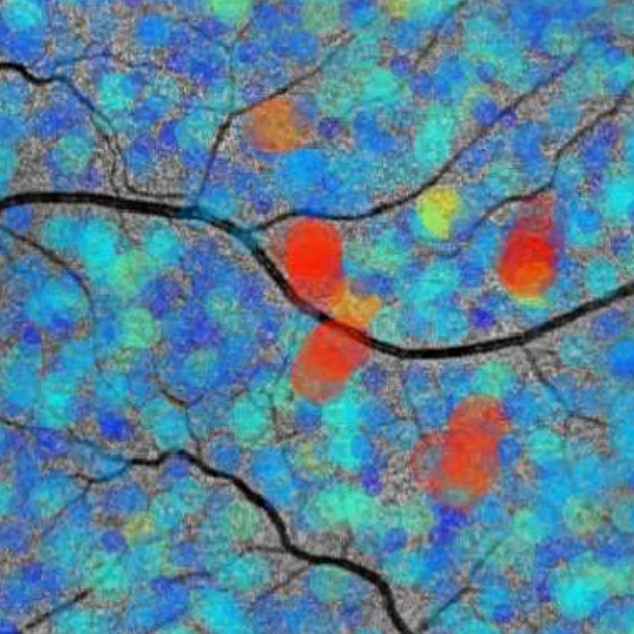}%
		\label{fig:SLOSampleImageResult}}
	\subfloat[(d) DiaretDB1]{\includegraphics[trim={6.6cm 4.3cm 13.3cm 11.1cm},clip,height=0.165\textwidth]{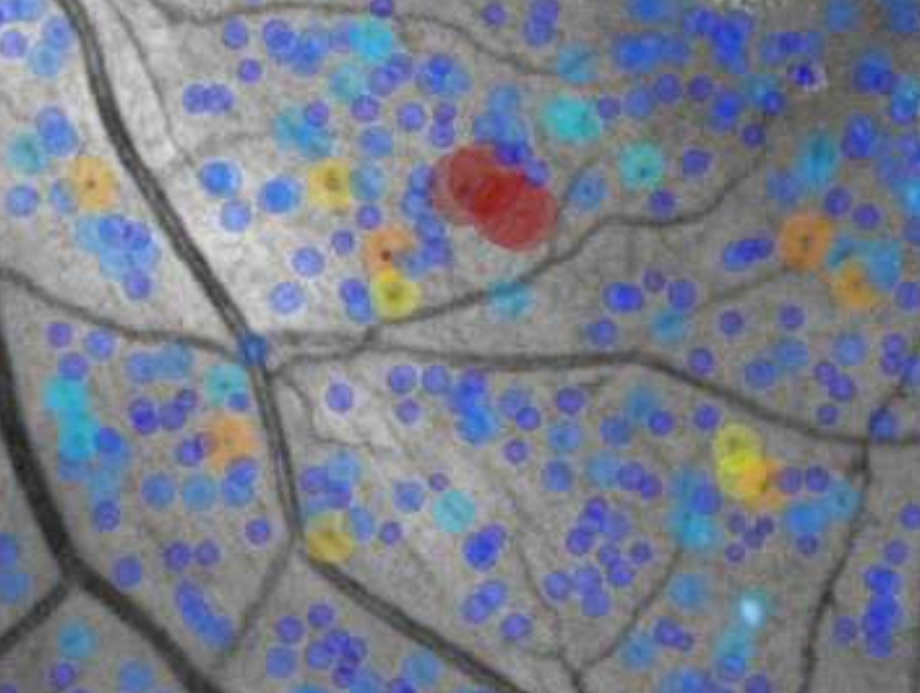}%
		\label{fig:Diretdb1SampleImageResult}}
	\subfloat[(e) ROC]{\includegraphics[trim={7cm 1.0cm 7cm 9.5cm},clip,height=0.165\textwidth]{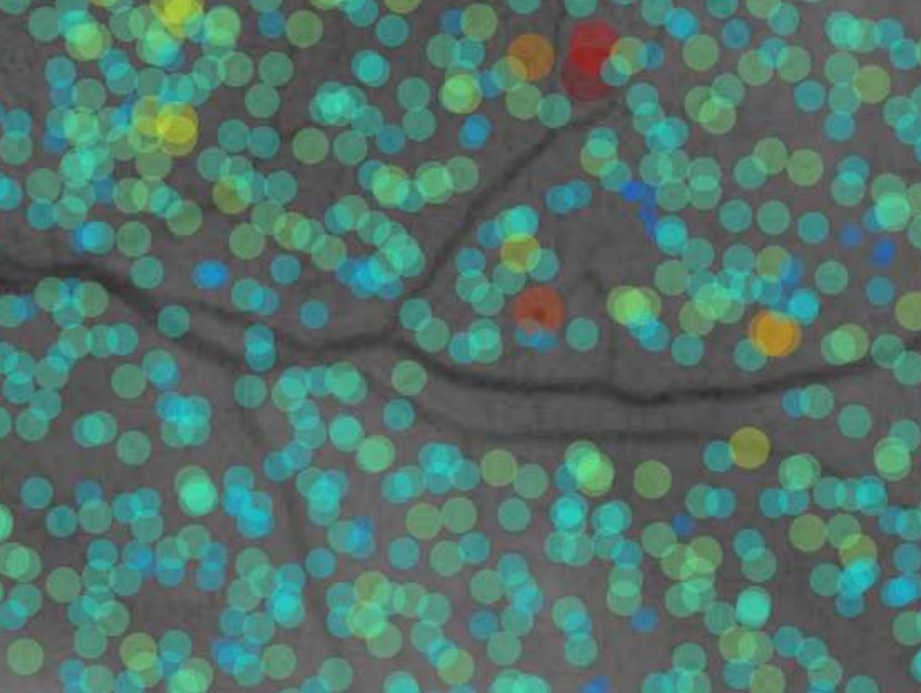}%
		\label{fig:ROCSampleImageResult}}
	\subfloat[]{\includegraphics[trim={0cm 0.20cm 0cm 0cm},clip,height=0.17\textwidth]{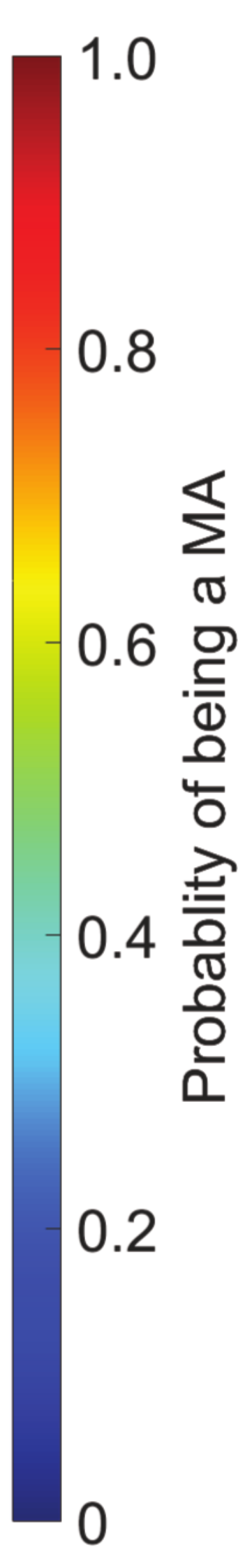}%
		\label{fig:colormap}}

		\vspace{-0.5em}
	\caption{Microaneurysms detection results by the proposed method on the five datasets;  \nth{1} row: green channel images in which the manually annotated MAs are shown by color-coded circles around them; yellow: only one expert; green: two experts agreement; magenta: three experts agreement; cyan: four experts agreement;\nth{2} row: the results of proposed method where the heat color map indicates the probabilities of being MA for the extracted candidates.}
	\label{fig:SampleImage}
	\vspace{-1.5em}
\end{figure*}

For the evaluation of proposed MA detection, we performed repeated 10-fold cross-validation for each dataset separately. In this approach, each dataset is divided randomly into ten equally sized partitions. Each partition is used as test data, while the other 9 partitions are used for training the classifier. The cross-validation procedure is repeated 10 times, yielding 10 performance results which are then averaged to produce a single estimation. Examples of MA detection on the five datasets are given in~\figurename{~\ref{fig:SampleImage}}.

To measure the performance of MA detection, we used the free-response operating characteristic (FROC) curve~\cite{Bunch1977} by plotting the sensitivity against the average number of false positives per image (FPI). Sensitivity represents the proportion of MAs correctly detected by the algorithm, while FPI is the number of non-MAs wrongly detected as MAs. For the sake of comparison with the other methods, the sensitivity values for the FPI rates values of $1/8$, $1/4$, $1/2$, $1$, $2$, $4$, and $8$ were obtained from the FROC curve. The final FROC score ($F_{score}$) is defined by the average of sensitivity values at these seven predefined FPIs~\cite{niemeijer2010retinopathy}.
In addition, we also obtained  the partial area under the FROC curves ($F_{AUC}$) between $1/8$ and $8$ FPI using trapezoidal integration and normalization by dividing with the maximum FPI~\cite{Antal2012}. 

\begin{figure*}[!h]
	\centering
	
	\subfloat[]{\includegraphics[trim={0cm 0cm 1cm 0cm},clip,height=0.28\textwidth]{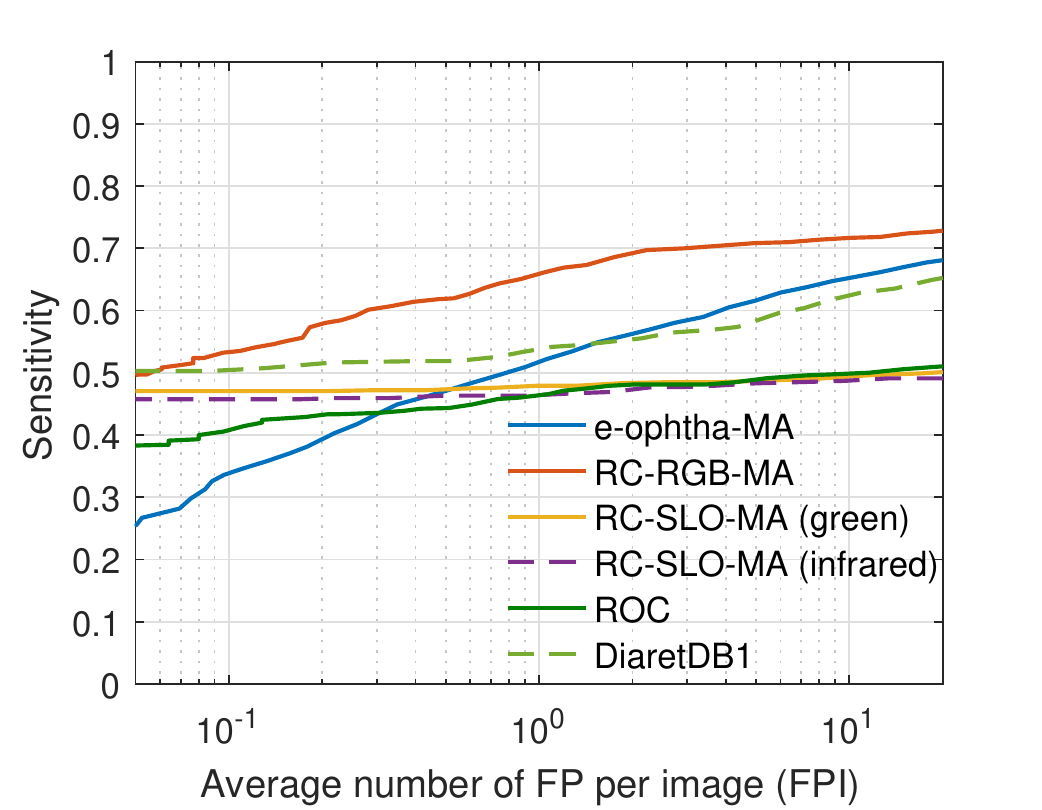}%
		\label{fig:FROCall}}
	\hfill
	\subfloat[]{\includegraphics[trim={0cm 0cm 1cm 0cm},clip,height=0.28\textwidth]{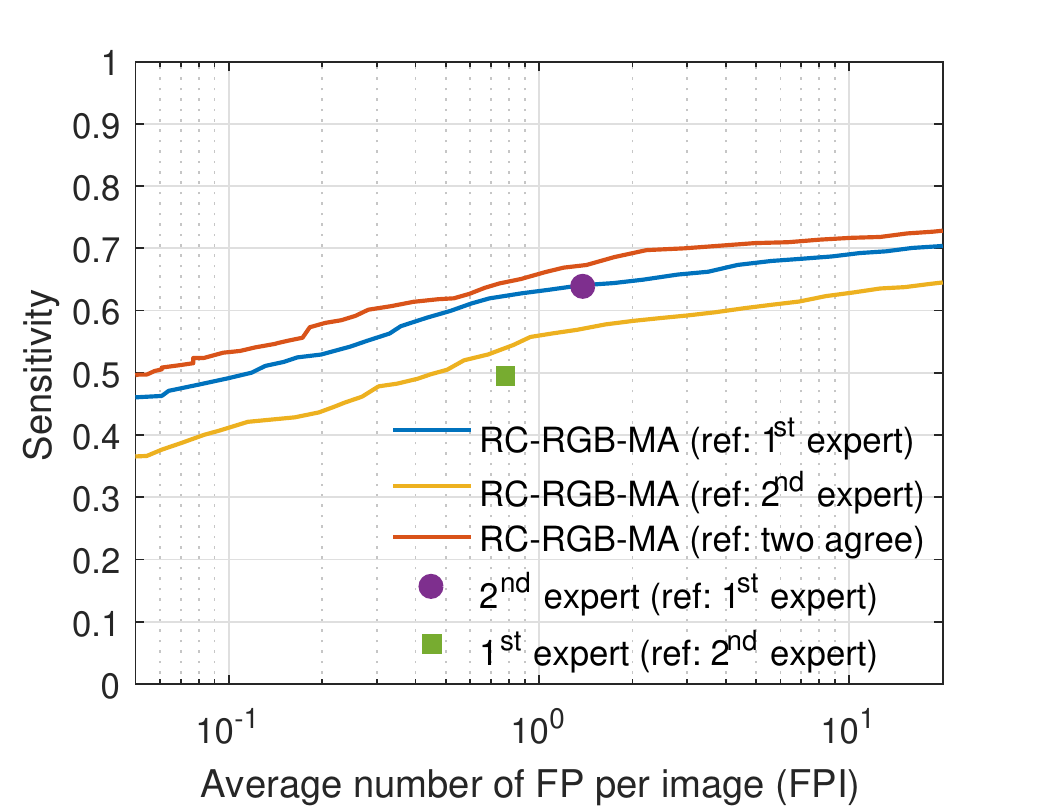}%
		\label{fig:FROC-RC-RGB-MA}}
	\hfill
	\subfloat[]{\includegraphics[trim={0cm 0cm 1cm 0cm},clip,height=0.28\textwidth]{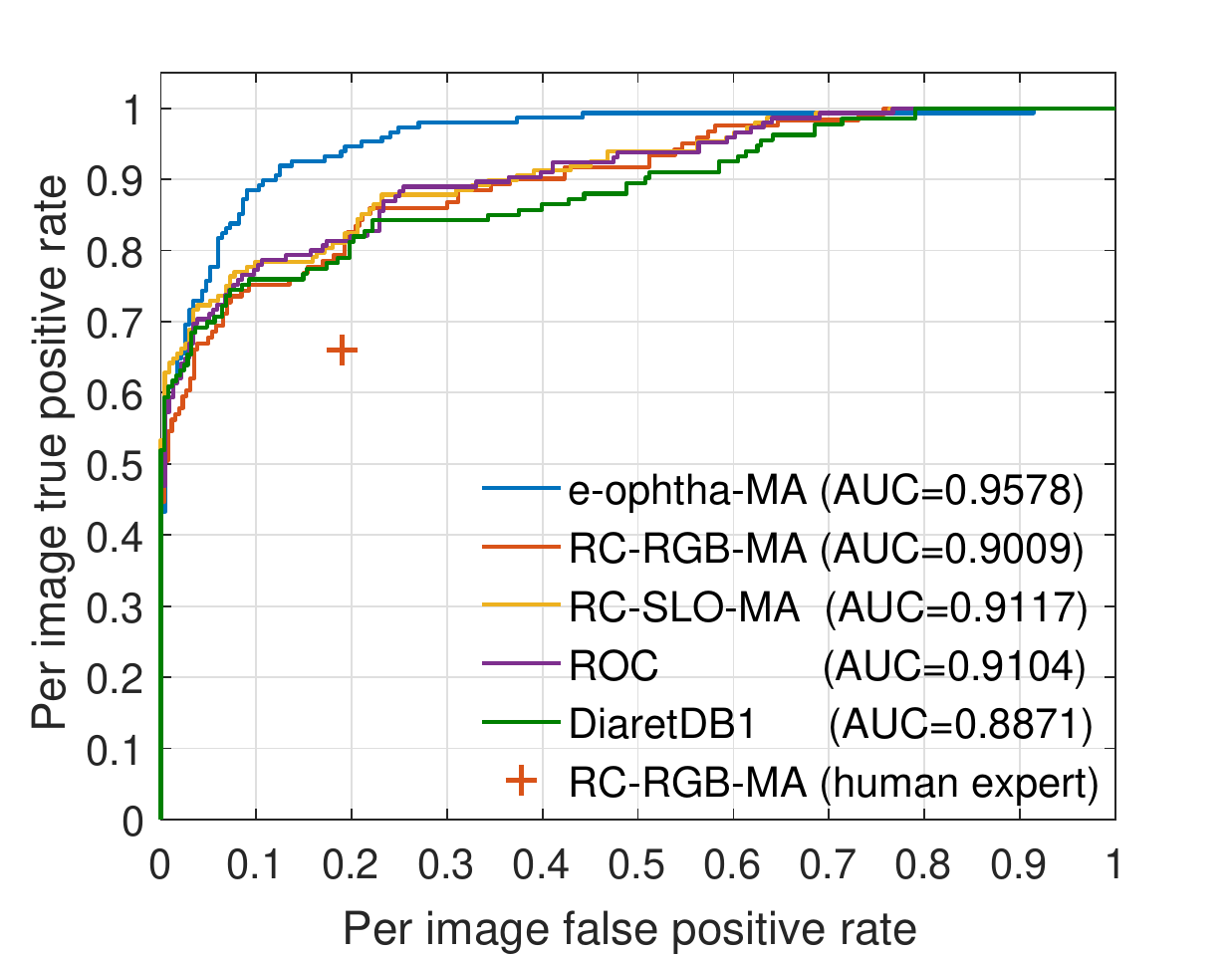}%
		\label{fig:ROC-subjects}}
	\vspace{-0.5em}
	\caption{(a) FROC curves of proposed MA detection method on 5 different datasets; (b) Comparison between FROC curves of proposed MA detection method with the performance of human experts on RC-RGB-MA; (c) ROC curves of image classification based on the highest score of detected MA on the 5 datasets.}
	\label{fig:FROC}
	\vspace{-1.5em}
\end{figure*}

The FROC curves of the proposed MA detection method on the five datasets are shown in \figurename{~\ref{fig:FROCall}}. Table~\ref{tab:AllDatasetsEvaluation} compares the $F_{score}$ and the $F_{AUC}$ of the proposed method with the state-of-the-art on the e-ophtha-MA, ROC and DiaretDB1 datasets. \figurename{~\ref{fig:FROC-RC-RGB-MA}} illustrates the results on the RC-RGB-MA dataset where the classifier is trained three times using the annotations provided by expert 1, expert 2 and the agreement of both experts, respectively. The performance of the two human experts are also demonstrated in this figure. 

\begin{table*}[htbp]
	\renewcommand{\arraystretch}{0.9}
	\centering
	\caption{Comparison of sensitivity values at predefined false positive per image rates for different MA detection methods using different datasets.}
	\vspace{-0.5em}
	{\footnotesize 
		\begin{tabular}{p{1.5cm}p{5cm}C{0cm}C{0.6cm}C{0.6cm}C{0.6cm}C{0.6cm}C{0.6cm}C{0.6cm}C{0.6cm}C{0.0cm}C{0.3cm}C{0.6cm}}
			\toprule
			\multirow{2}[0]{*}{Dataset}&\multirow{2}[0]{*}{Method} &  & \multicolumn{8}{c}{Sensitivity against FPI}       & \multicolumn{1}{c}{\multirow{2}[0]{*}{$F_{score}$}} & \multicolumn{1}{l}{\multirow{2}[0]{*}{$F_{AUC}$}}  \\
			\cline{4-10}
			\noalign{\vskip 1mm}   
			&\multicolumn{1}{c}{} & \multicolumn{1}{c}{}   & $1/8$ & $1/4$ & $1/2$ & $1$ & $2$ & $4$ & $8$ &  &       &  \\
			\midrule
			\midrule
			\parbox[t]{2mm}{\multirow{3}{*}{\rotatebox[origin=c]{0}{e-ophtha-MA}}} 
			& Proposed method 							&& \textbf{0.358} & \textbf{0.417} & \textbf{0.471} & \textbf{0.522} & \textbf{0.558} & 0.605 & 0.638 && \textbf{0.510}		& 0.575   \\
			& Wu \textit{et al.} (2017)~\cite{Wu2017106}		&& 0.063 & 0.117 & 0.172 & 0.245 & 0.323 & 0.417 & 0.573 && 0.273		& 0.386\\
			& Zhang (2014)~\cite{Zhang2014}						&& 0.170 & 0.240 & 0.320 & 0.440 & 0.540 & \textbf{0.630} & \textbf{0.740} && 0.440		&\textbf{0.586}\\
			\cline{1-13}\noalign{\vskip 1mm}  
			\parbox[t]{2mm}{\multirow{3}{*}{\rotatebox[origin=c]{0}{RC-RGB-MA}}} 
			& Proposed method (1st expert)	&& 0.511 & 0.542 & 0.599 & 0.633 & 0.650 & 0.673 & 0.687 && 0.614 	& 0.650   \\
			& Proposed method (2nd expert) 	&& 0.421 & 0.452 & 0.505 & 0.558 & 0.584 & 0.597 & 0.623 && 0.534 	& 0.579  \\
			& Proposed method (agreement) 	&& \textbf{0.541} & \textbf{0.591} & \textbf{0.618} & \textbf{0.662} & \textbf{0.697} & 0\textbf{.704} & \textbf{0.714} && \textbf{0.647 }	& \textbf{0.682}   \\
			
			\cline{1-13}\noalign{\vskip 1mm} 
			\parbox[t]{2mm}{\multirow{2}{*}{\rotatebox[origin=c]{0}{RC-SLO-MA }}} 
			& Proposed method (green)						& & \textbf{0.471 }& \textbf{0.472} & \textbf{0.472} & \textbf{0.479} & \textbf{0.483} & \textbf{0.485} & \textbf{0.491} && \textbf{0.479} 	& \textbf{0.477 } \\
			&Proposed method (infrared)	& & 0.457 & 0.459 & 0.463 & 0.463 & 0.469 & 0.479 & 0.485 && 0.468 	& 0.468  \\
			
			\cline{1-13}\noalign{\vskip 1mm} 
			\parbox[t]{2mm}{\multirow{16}{*}{\rotatebox[origin=c]{0}{ROC}}} 
			& Proposed method 										&& \textbf{0.435} & \textbf{0.443} & \textbf{0.454} & 0.476 & 0.481 & 0.495 & 0.506 && \textbf{0.471} 	& 0.484	\\ 	
			& Wang \textit{et al.} (2017)~\cite{Wang2017}			&& 0.273 & 0.379 & 0.398 & \textbf{0.481}& \textbf{0.545} & 0.576 & 0.598 && 0.464 	& 0.543   \\
			& Wu \textit{et al.} (2017)~\cite{Wu2017106}			&& 0.037 & 0.056 & 0.103 & 0.206 & 0.295 & 0.339 & 0.376 && 0.202	& 0.302\\
			& Seoud \textit{et al.} (2016)~\cite{Seoud2016}			&& 0.212$^\dagger$ & 0.318$^\dagger$ & 0.359$^\dagger$ & 0.410$^\dagger$ & 0.470$^\dagger$ & 0.533$^\dagger$ & 0.609$^\dagger$ && 	0.420	& 0.505$^\dagger$	\\
			& Dai \textit{et al.} (2016)~\cite{Dai2016}						&& 0.219 & 0.257 & 0.338 & 0.429 & 0.528 & 0.598 & \textbf{0.662} && 0.433 	& \textbf{0.553}  \\
			& Adal \textit{et al.} (2014)~\cite{Adal20141}					&& 0.204 & 0.255 & 0.297 & 0.364 & 0.417 & 0.478 & 0.532 && 0.364	& 0.446   \\
			& Pereira \textit{et al.} (2014) ~\cite{Pereira2014179}			&& 0.053 & 0.083 & 0.135 & 0.187 & 0.276 & 0.407 & 0.540 && 0.240  	& 0.366   \\
			& Lazar \textit{et al.} (2013)~\cite{Lazar2013} 				&& 0.251 & 0.312 & 0.350 & 0.417 & 0.472 & 0.542 & 0.615 && 0.423 	& 0.510   \\
   		& DRSCREEN: Antal \textit{et al.} (2012)~\cite{Antal2012}			&& 0.173 & 0.275 & 0.380 & 0.444 & 0.526 & \textbf{0.599 }& 0.643 && 0.434 	& 0.551   \\
			& Fegyver \textit{et al.} (2012)~\cite{Fegyver2012}				&& 0.248 & 0.309 & 0.341 & 0.417 & 0.487 & 0.554 & 0.601 && 0.422 	& 0.514  \\
			& OKmedical II: \textit{Zhang et al.} (2012)~\cite{Zhang201278}	&& 0.175 & 0.242 & 0.297 & 0.370 & 0.437 & 0.493 & 0.569 && 0.369	& 0.465   \\
			& ISMV: Giancardo \textit{et al.} (2011)~\cite{Giancardo2011}	&& 0.217 & 0.270 & 0.366 & 0.407 & 0.440 & 0.459 & 0.468 && 0.375 	& 0.435  \\
			& IRIA Group (2011)~\cite{Ram2011}								&& 0.041 & 0.160 & 0.192 & 0.242 & 0.321 & 0.397 & 0.493 && 0.264 	& 0.368   \\
			& OKmedical: Zhang \textit{et al.} (2010)~\cite{Zhang20102237}	&& 0.198 & 0.265 & 0.315 & 0.356 & 0.394 & 0.466 & 0.501 && 0.357	& 0.430   \\
			& GIB: Sanchez \textit{et al.} (2009)~\cite{Sanchez2009} 		&& 0.190 & 0.216 & 0.254 & 0.300 & 0.364 & 0.411 & 0.519 && 0.322 	& 0.399   \\
			& Fujita Lab (2009)~\cite{Mizutani2009}							&& 0.181 & 0.224 & 0.259 & 0.289 & 0.347 & 0.402 & 0.466 && 0.310  	& 0.378   \\
			& LaTIM: Quellec \textit{et al.} (2008)~\cite{Quellec2008}		&& 0.166 & 0.230 & 0.318 & 0.385 & 0.434 & 0.534 & 0.598 && 0.381 	& 0.489  \\
			& Waikato group (2008)~\cite{Cree2008}  						&& 0.055 & 0.111 & 0.184 & 0.213 & 0.251 & 0.300 & 0.329 && 0.206 	& 0.273   \\
			& Niemeijer \textit{et al.} (2005)~\cite{Niemeijer2005}			&& 0.243 & 0.297 & 0.336 & 0.397 & 0.454 & 0.498 & 0.542 && 0.395 	& 0.469  \\
			
			\cline{1-13}\noalign{\vskip 1mm} 
			\parbox[t]{2mm}{\multirow{3}{*}{\rotatebox[origin=c]{0}{DiaretDB1}}} 
			& Proposed method 									&& \textbf{0.507} & \textbf{0.517} & \textbf{0.519} & \textbf{0.542} & \textbf{0.555} & 0.574 & 0.617 && \textbf{0.547} 	& \textbf{0.565} \\
			& Seoud \textit{et al.} (2016)~\cite{Seoud2016}		&&0.140$^\dagger$ & 0.175$^\dagger$ & 0.250$^\dagger$ & 0.323$^\dagger$ & 0.440$^\dagger$ & 0.546$^\dagger$ & 0.642$^\dagger$ && 0.354	& 0.495$^\dagger$	\\
			& Dai \textit{et al.} (2016)~\cite{Dai2016}			&& 0.035 & 0.058 & 0.112 & 0.254 & 0.427 & \textbf{0.607} & \textbf{0.755} && 0.321	& 0.527	\\
			& Adal \textit{et al.} (2014)~\cite{Adal20141} 		&& 0.024$^\dagger$ & 0.033$^\dagger$ & 0.045$^\dagger$ & 0.103$^\dagger$ & 0.204$^\dagger$ & 0.305$^\dagger$ & 0.571$^\dagger$ && 0.184$^\dagger$	& 0.308$^\dagger$ 	\\
			& DRSCREEN (2012)~\cite{Antal2012}					&& 0.001 & 0.003 & 0.009 & 0.020 & 0.059 & 0.140 & 0.257 && 0.070	& 0.130\\
			\bottomrule
 			\noalign{\vskip 1mm}
			\multicolumn{13}{l}{FPI: number of false positives per image; $F_{score}$: average of sensitivities at different FPIs; $F_{AUC}$: partial area under FROC curve; $Acc$: accuracy.}\\
			\multicolumn{13}{l}{\textbf{Bold} value indicates highest numbers for each dataset;$^\dagger$ values are extracted from the plots reported in~\cite{Seoud2016,Adal20141} using WebPlotDigitizer application~\cite{WebPlotDigitizer}.}\\ 
		\end{tabular}%
	}
	\label{tab:AllDatasetsEvaluation}%
	\vspace{-.5em}
\end{table*}%

\subsection{Image Classification Evaluation}
The proposed MA detection method extracts several candidates per image and assigns a probability value to each candidate. For the image classification, the highest probability among all candidates is considered as the image score. If the image score is greater than a certain threshold level, it will indicate that the image has signs of diabetic retinopathy and contains at least one detected MA. A low image score shows the absence of MAs indicating a healthy retina. For the valuation, the area under the ROC curve (AUC) is obtained for all five datasets as shown in \figurename{~\ref{fig:ROC-subjects}} in which the human expert performance on the RC-RGB-MA dataset is also demonstrated by a red plus sign.   


\subsection{Features Importance Analysis}

\begin{figure*}[!h]
	\centering
	
	\subfloat[]{\includegraphics[trim={1.1cm 0cm 1.1cm 0cm},clip,width=0.30\textwidth]{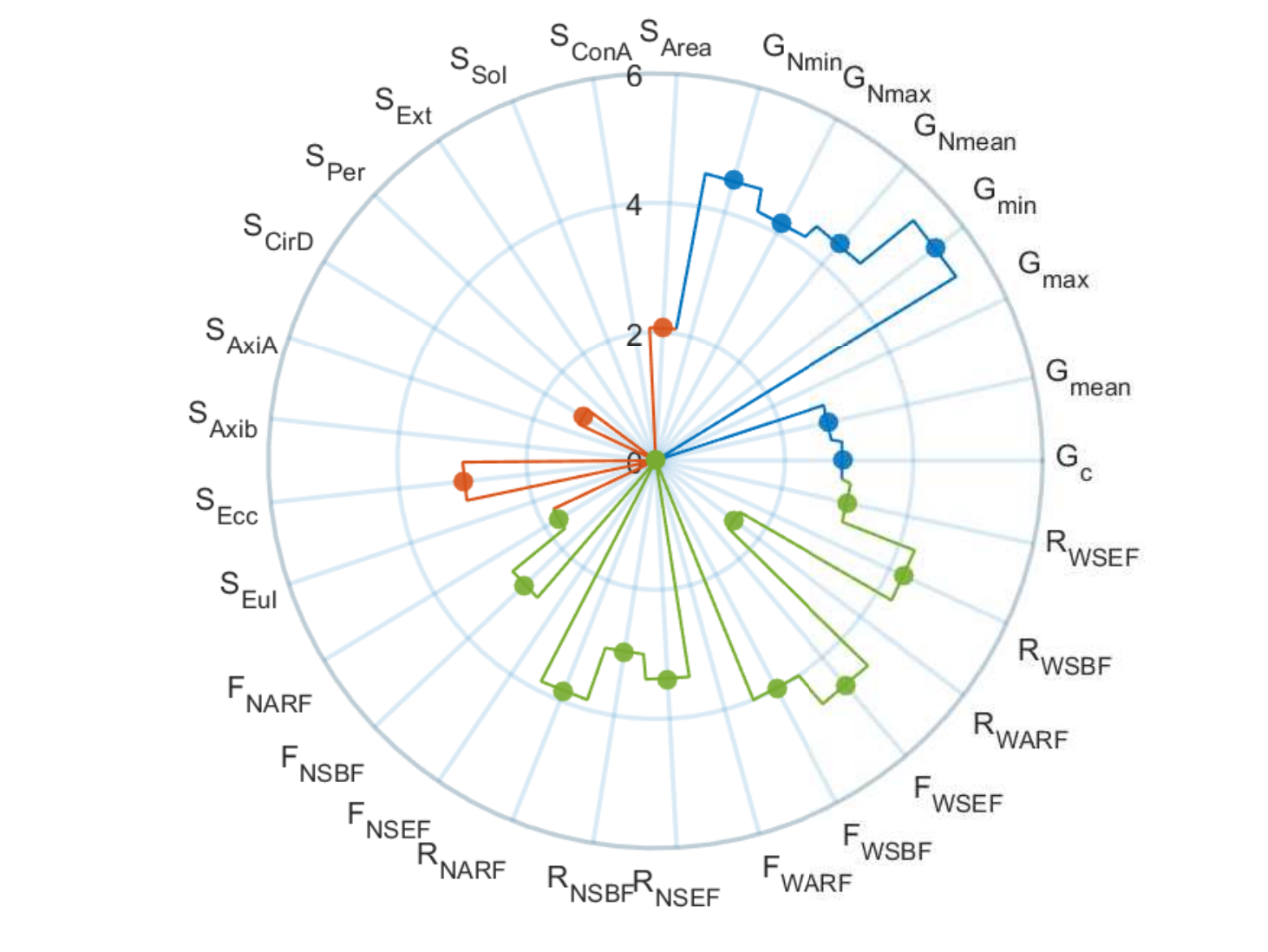}%
		\label{fig:FeaturesImportanceGini}}
	\hfill
	\subfloat[]{\includegraphics[trim={1.1cm 0cm 1.1cm 0cm},clip,width=0.30\textwidth]{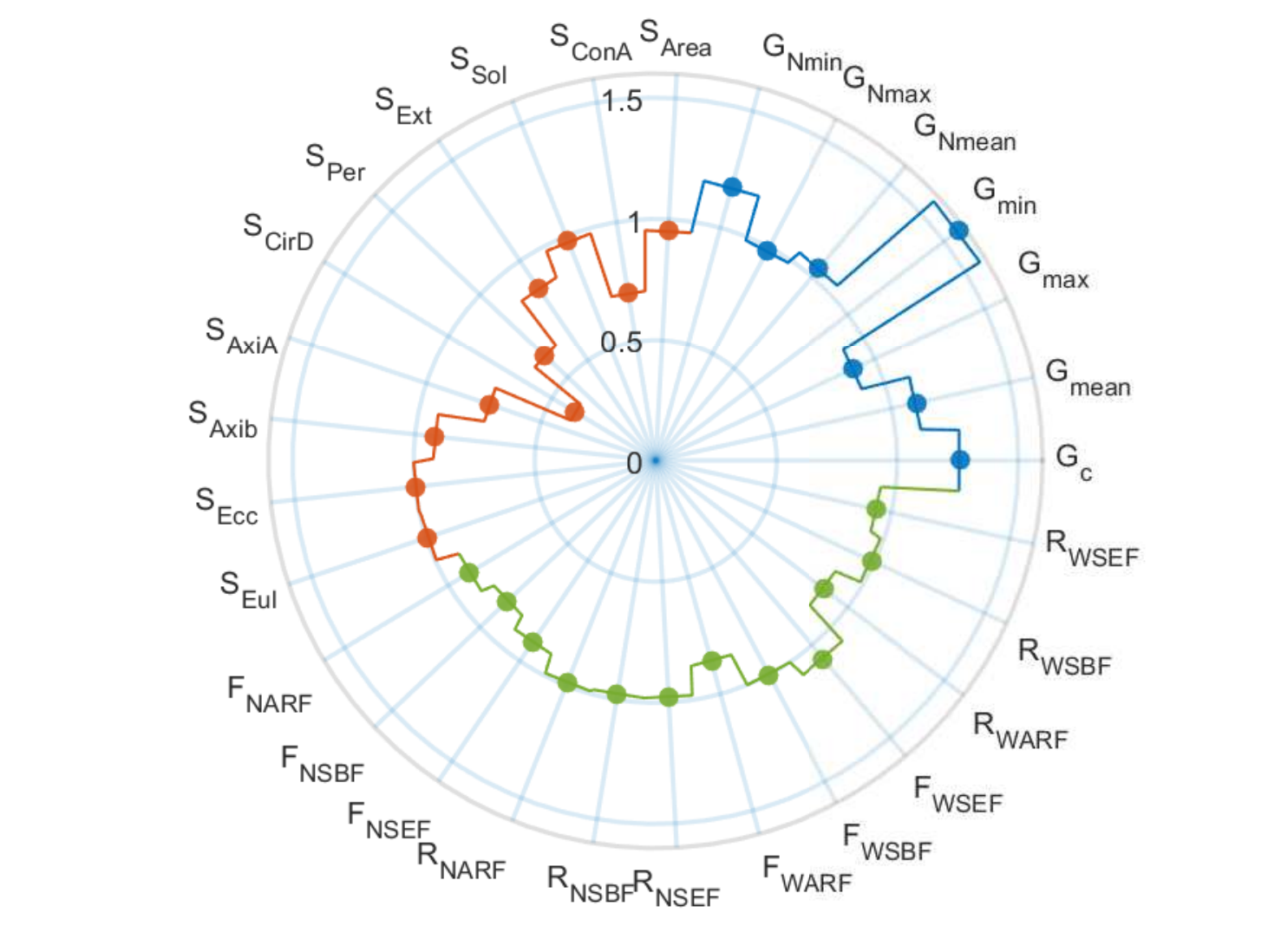}%
		\label{fig:FeaturesImportanceDeviance}}
	\hfill
	\subfloat[]{\includegraphics[trim={1.1cm 0cm 1.1cm 0cm},clip,width=0.30\textwidth]{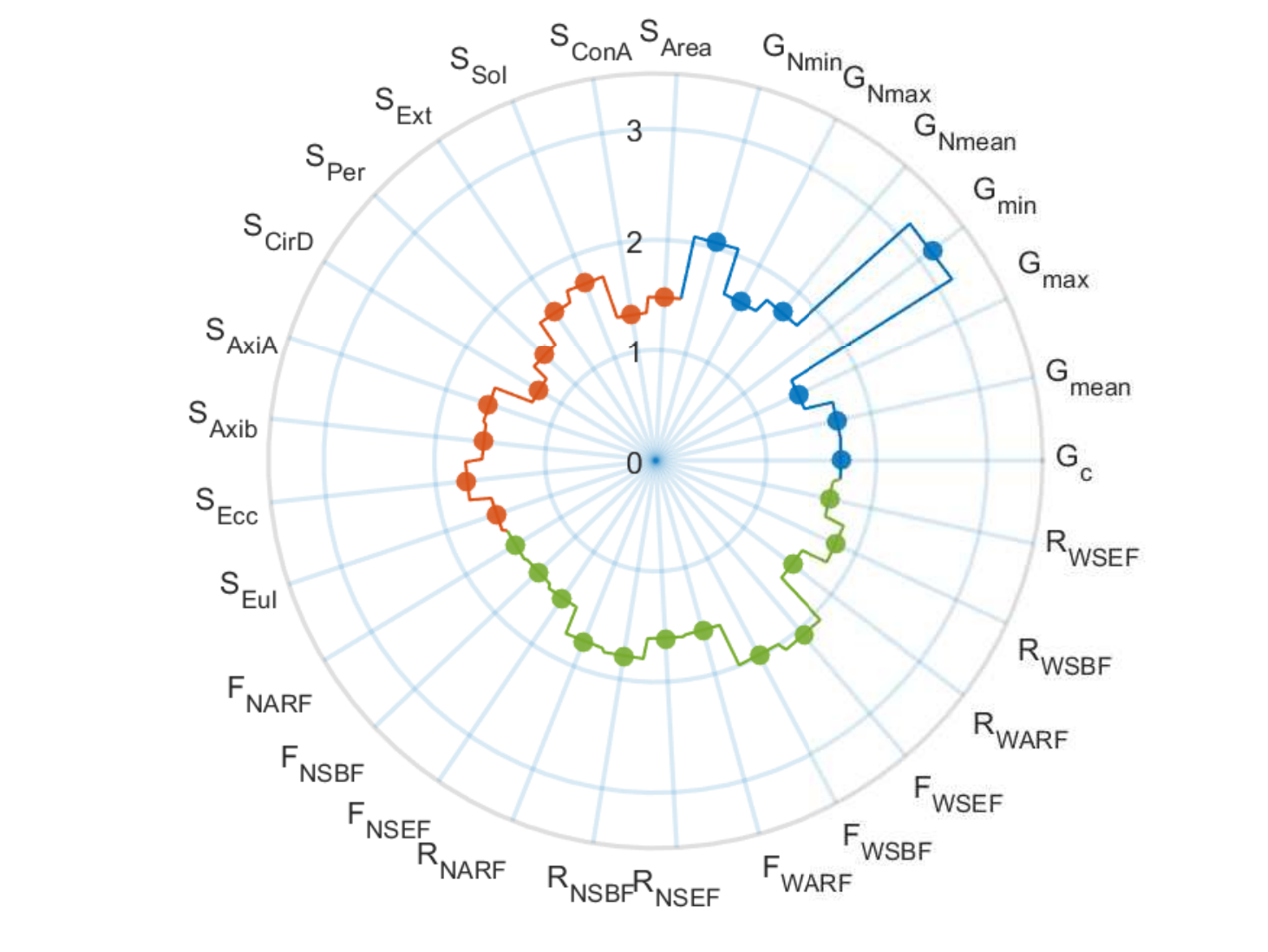}%
		\label{fig:FeaturesImportanceTwoing}}
	\vspace{-0.5em}
	\caption{Features importance analysis by calculating the impurity of nodes using different split criteria (a)  Gini's diversity index, (b) Deviance and (c) Twoing rule. For visualization purpose, all the importance values are plotted using the logarithmic scale in base 10. Blue color represents the intensity-based features while the shape-based and LCF-based features are demonstrated by orange and green colors, respectively.}
	\label{fig:FeatureImportance}
	\vspace{-1.5em}
\end{figure*}
In the section, we investigate the importance of the extracted features to show the relative contribution of different types of features for the MAs classification. The feature importance analysis for the RUSBoost classifier is performed by summing changes in the risk due to splits on every predictor and dividing the sum by the number of branch nodes. This sum is taken over best splits found at each branch node. 
The feature importance associated with this split is computed as the difference between the risk for the parent node and the total risk for the two children. The nodes are split based on the impurity which is dependent to the split criteria. Here we calculated the importance values using the following split criterion:  

\begin{enumerate}
	\item Gini's Diversity Index~\cite{breiman1996technical}:  $1-\sum_{i=1}^{2} p^2(i)$, \\
	where $p(i)$ is the observed fraction of classes with class $i$ that reach the node. A pure node (with just one class) has Gini index 0; otherwise the Gini index is positive. 
	\item Deviance~\cite{breiman1996technical}: $ -\sum_{i=1}^{2}p(i)\log p(i)$,\\
	where $p(i)$ is defined the same as for the Gini index, and a pure node has deviance 0; otherwise, it is positive.
	\item Twoing rule~\cite{breiman1996technical}: $ P(L)P(R)\left ( \sum_{i=1}^{2}\left | L(i)-R(i) \right | \right )^2$,\\
	where $L(i)$ denotes the fraction of members of class $i$ in the left child node after a split, and $R(i)$ denotes the fraction of members of class $i$ in the right child node after a split. $P(L)$ and $P(R)$ are the fractions of observations that split to the left and right respectively.
	If the expression is large, the split made each child node purer. Similarly, if the expression is small, the split made each child node more similar to each other, and hence more similar to the parent node, and so the split did not increase node purity.
\end{enumerate}

We evaluated all the 29 features using the ‘Gini's Diversity Index’, ‘Deviance’ and ‘Twoing rule’ on a subset of images from the e-ophtha-MA dataset. The obtained feature importance maps are shown in \figurename{~\ref{fig:FeatureImportance}} and the corresponding description of each feature is given in Table{~\ref{tab:featuresList}}.
The features with the top 12 maximum importance values in \figurename{~\ref{fig:FeatureImportance}} are selected as the reduced subsets, which are then used to train the RUSBoost classifier.  
The intensity-based features and the proposed local convergence filter-based features are frequently used in decision making for all the three split criteria, while the shape-based features have the least contributions.

\begin{figure}[!h]
	\centering
	
\includegraphics[trim={0cm 0cm 0cm 0.5cm},clip,width=0.47\textwidth]{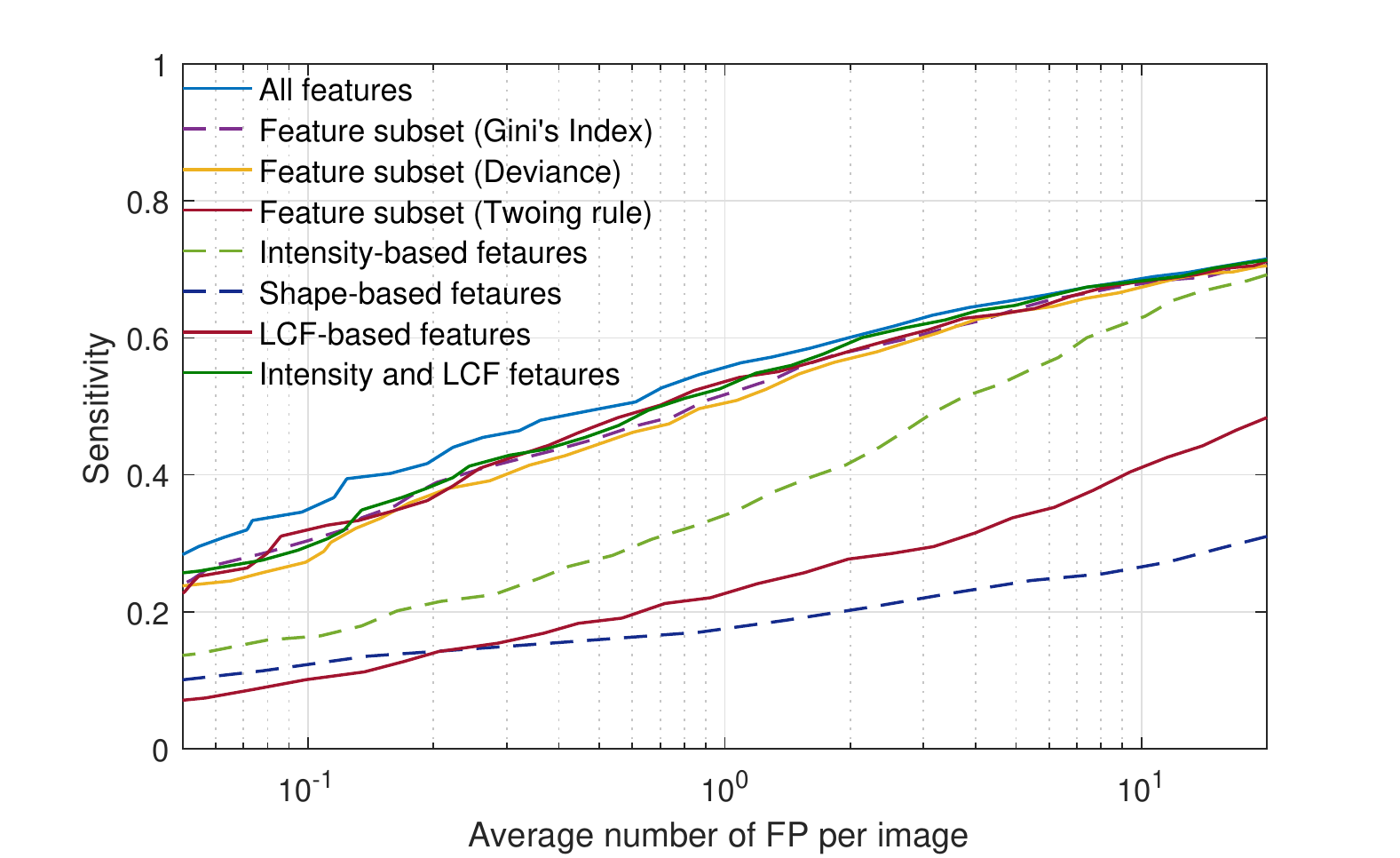}%
	\caption{FROC curves of the proposed method using different sets of features on a subset of images from the e-ophtha-MA dataset (without taking into account the missed candidates in the candidate extraction step).}
	\label{fig:FeatureSubsetFROC}
	\vspace{-1.5em}
\end{figure}

We also trained the classifier using each category of features individually or combined with each other. The results are shown in Table~\ref{tab:FeatureImportance} and compared with the performance of using the full feature set.  The values reported in this table are only based on the performance of trained classifiers without including the error of the candidate extraction step.
Although the intensity-based features have the most substantial contributions ($F_{score}=0.366$ and $F_{AUC} = 0.473$) compared to two other feature categories (see~\figurename{~\ref{fig:FeatureSubsetFROC}}), inclusion of the LCF-based features improves the performance significantly, resulting in a $F_{score}$ of 0.518 and a $F_{AUC}$ of 0.604. On the other hand, including shape-based features improves the performance only slightly.
As illustrated in \figurename{~\ref{fig:FeatureImportance}}, the most important descriptor in the category of intensity-based features is the minimum green intensity ($G_{min}$), and among the LCF-based features the response of the SEF filter on the gradient-weighted image ($F_{WSEF}$) has the highest contribution. 


%
%

%

\begin{table*}[htbp]
	\renewcommand{\arraystretch}{0.9}
	\centering
	\caption{Classifier performance comparison using different sets of features on a subset of images from the e-ophtha-MA dataset.}
	\vspace{-0.5em}
	{\footnotesize 
	\begin{tabular}{p{6.7cm}C{0cm}C{0.6cm}C{0.6cm}C{0.6cm}C{0.6cm}C{0.6cm}C{0.6cm}C{0.6cm}C{0.0cm}C{0.3cm}C{0.6cm}}
		\toprule
		\multirow{2}[0]{*}{Features subset} &  & \multicolumn{7}{c}{Sensitivity against FPI}       && \multicolumn{1}{c}{\multirow{2}[0]{*}{$F_{score}$}}& \multicolumn{1}{c}{\multirow{2}[0]{*}{$F_{AUC}$}}  \\
		\cline{3-9}
		\noalign{\vskip 1mm}   
		\multicolumn{1}{c}{} & \multicolumn{1}{c}{}   & $1/8$ & $1/4$ & $1/2$ & $1$ & $2$ & $4$ & $8$ &  &       &  \\
		\midrule
		\midrule
		All features ($29: f_{1-29}$) &		     											& \textbf{0.394} & \textbf{0.455} & \textbf{0.494} & \textbf{0.564} & \textbf{0.600} & \textbf{0.644} & 0.673 && \textbf{0.546}      & \textbf{0.612}  \\
		Gini's index   ($12: f_{2}, f_{4-7},f_{13},f_{21-22},f_{25-26},f_{28-29}$) &		& 0.321 & 0.410 & 0.453 & 0.524 & 0.575 & 0.625 & 0.673 && 0.512        & 0.593 \\
		Deviance  ($12: f_{1-2},f_{4-5},f_{7},f_{17},f_{21-22},f_{25-26},f_{28-29}$) &      & 0.322 & 0.391 & 0.443 & 0.509 & 0.564 & 0.627 & 0.657 && 0.502       & 0.585  \\
		Twoing rule  ($12: f_{2},f_{4-5},f_{7},f_{10},f_{21-22},f_{24-26},f_{28-29}$) &     & 0.333 & 0.410 & 0.484 & 0.542 & 0.582 & 0.628 & 0.673 && 0.522        & 0.599  \\
		Only intensity-based features ($7: f_{1-7}$) &       								& 0.179 & 0.225 & 0.282 & 0.346 & 0.414 & 0.514 & 0.600 && 0.366        & 0.473  \\
		Only shape-based features ($10: f_{8-17}$) &       									& 0.135 & 0.149 & 0.159 & 0.170 & 0.207 & 0.227 & 0.255 && 0.186        & 0.216   \\
		Only LCF-based features    ($12: f_{18-29}$) &       								& 0.112 & 0.154 & 0.183 & 0.220 & 0.277 & 0.315 & 0.378 && 0.234        & 0.298  \\
		Intensity and LCF-based features  ($19: f_{1-7},f_{18-29}$) &       				& 0.319 & 0.413 & 0.455 & 0.525 & \textbf{0.600} & 0.640 & \textbf{0.674} && 0.518        & 0.604  \\
		\bottomrule
		\noalign{\vskip 1mm}
		\multicolumn{12}{l}{$F_{score}$: average of sensitivities at different FPIs; $F_{AUC}$: partial area under FROC curve; $Acc$: accuracy; \textbf{Bold} value indicates highest numbers.}  \\
	\end{tabular}%
	}
	\label{tab:FeatureImportance}%
	\vspace{-1.5em}
\end{table*}%


\section{Discussion and Conclusion}\label{sec:Discussion}
The proposed method is in the form of a pipeline of techniques where the performance of each step depends on the output of the previous step. Hence, the method is validated extensively in each one of three stages: candidate extraction, MA detection, and image classification. 

\subsection{Candidate Extraction Performance}
The goal of the candidate extraction is to reduce the computational burden by decreasing the number of objects for further analysis in the next step. 
As shown in Table~\ref{tab:CandidateEvaluation}, the proposed candidate extractor outperforms the state-of-the-art methods by achieving a sensitivity value of 0.82 on the ROC dataset.
The average of false positives per image is higher than the ones reported by other methods, however, it is less than $0.06\%$ of the total number of pixels in one image. 
In this step, it is important to include true MAs as much as possible without considering the balance between sensitivity and specificity. 
The false positives are later discarded in the classification step.
The evaluation on other datasets demonstrates that the introduced candidate extractor performs better on the e-ophtha-MA ($Sen = 0.95$) and the RC-RGB-MA ($Sen = 0.94$) since both datasets contain high resolution and high quality images.

\subsection{Microaneurysm Detection Performance}
The detection of MAs highly depends on the imaging device characteristics and image properties such as resolution, modality, compression technique, illumination and contrast variation.
For this reason, we evaluated the proposed method on five datasets. The results are compared with the state-of-the-arts on the e-ophtha-MA, ROC and DiaretDB1 datasets, while the RC-RGB-MA and  RC-SLO-MA datasets are used for the comparison with human experts and for the validation on a different image modality (SLO images).

On the high quality images of e-ophtha-MA dataset our method achieves a $F_{score}$ of 0.510 and a $F_{AUC}$ of 0.575 which are significantly higher than the values reported by Wu \textit{et al.}~\cite{Wu2017106}.
On the same dataset Zhang~\cite{Zhang2014} used contextual descriptors for the classification of MAs with $F_{score}$ equal to 0.440 and with a slightly better partial AUC of 0.586.  

\figurename{~\ref{fig:FROC-RC-RGB-MA}} and Table~\ref{tab:AllDatasetsEvaluation} show the performance of the proposed method on RC-RGB-MA, as well as the inter-expert variability between two human experts. The $F_{score}$ values of 0.534 and 0.614 are obtained by using the annotations provided by the \nth{1}~expert  and \nth{2}~expert respectively while using the agreement of two experts on the annotations results in a better performance ($F_{score}= 0.647$). In all three cases of different training approaches, the proposed method has a similar or slightly better performance than the human experts.

The RC-SLO-MA dataset provides a unique possibility for the validation of the proposed method on a different image modality (SLO). The green laser usually provides a better contrast between vessels and background, while the infrared laser penetrates deeper into the retina and visualizes pathologies in different layers of the retina. 
The proposed method performs slightly better on the green channel than the infrared image while the obtained $F_{score}$ values for both channels are relatively lower than the values achieved on the RGB datasets since the SLO images typically contain more background noise than conventional fundus images (RGB) and have a lower spatial resolution.

The images of the ROC dataset were acquired at different resolutions using different cameras. The variability in image resolution as well as the presence of noise and artifacts makes it more challenging to detect MAs in the images of this dataset. However, our method overcomes these difficulties and achieves a $F_{score}$ of 0.471 outperforming the state-of-the-art approaches.  The sensitivity values at predefined FPIs are compared with the others in Table~\ref{tab:AllDatasetsEvaluation} showing our method has higher sensitivities at FPI values of 1/8, 1/4 and 1/2. Wang \textit{et al.}~\cite{Wang2017} achieved slightly higher sensitivity at 1 FPI. Although some methods~\cite{wang2014hierarchical,Dai2016,Seoud2016,Lazar2013,Antal2012} reported higher sensitivities at 2, 4 and 8 FPI, it should be noted that the FPI of 1.08 is considered as an indication of ``clinically acceptable" FPI and higher values are not adequate for a computer-aided diagnosis system in clinical environments~\cite{niemeijer2010retinopathy}.

Although most of the images in DiaretDB1 dataset contain several dark spots and artefacts, our proposed method achieves remarkably better results compared to other methods. As shown in Table~\ref{tab:AllDatasetsEvaluation}, our method achieves a $F_{score}$ value of 0.547 and  a $F_{AUC}$ of 0.564 for the DiaretDB1 dataset. These values are significantly higher than the ones based on the dynamic shape features presented by Seoud \textit{et al.}~\cite{Seoud2016} and the gradient vector analysis technique proposed by Dai \textit{et al.}~\cite{Dai2016}.

\subsection{Image Classification Performance}
\figurename{~\ref{fig:FROCall}} illustrates the ROC curves of image classification, where the highest probability value among all candidates is assigned to  the image as its score. The highest AUC is obtained for the e-ophtha-MA dataset ($\text{AUC}=0.96$) while the lowest one is for DiaretDB1 dataset ($\text{AUC}=0.89$). The dark spot artefacts in the DiaretDB1 have high probability values which can cause a wrong image classification resulting in a lower AUC. Although the MAs detection on the RC-SLO-MA dataset has a lower performance than the other datasets, the high AUC value of 0.91 is obtained for the image classification. Visual inspection, as well as the high AUC, reveals that the obvious MAs are much easier to detect in SLO images compared to RGB images. The comparison between the ROC curve of our method on the RC-RGB-MA dataset ($\text{AUC} = 0.90$) and the sensitivity/specificity of the \nth{2} expert (reference: \nth{1} expert ) demonstrates the relatively high success rate of the proposed method in DR detection (see \figurename{~\ref{fig:FROCall}}).

%

\subsection{Computation Time}
The fully automated MAs detection method is developed in MATLAB 2016b (MathWorks, Inc.) with an average computation time of 3 minutes per image using an Intel Core i7-5820 CPU at 3.30 GHz. 

\subsection{Conclusion}\label{sec:Conclusion}
In this paper, we have proposed a new method for detecting MAs in retinal images using a gradient weighting technique, a new set of features based on local convergence filters (LCF) and a random undersamping boosting classifier. Feature importance analysis demonstrates that LCF-based descriptors can well characterize the low contrast MAs since the LCF filters are based on gradient convergence and not intensity. The performance of the proposed method on the ROC, DiaretDB1 and e-ophtha-MA datasets shows the competitiveness of the introduced approach against state-of-the-art techniques. Moreover, the evaluation results on five public datasets demonstrate that the proposed MAs detection method is insensitive to the characteristics of the imaging device, image resolution and image modality. Future work will involve exploiting the introduced LCS-based features for the detection of dot hemorrhages and bright lesions namely exudates. 
\section*{Acknowledgment}
The authors would like to thank He University Eye Hospital, Shenyang, China for providing color fundus images and Shanshan Zhu, Estera Ana Zarnescu, Anamaria Carla Vass, Mihaela Giurgia and Hans de Ferrante for their help with RC-RGB-MA and RC-SLO-MA datasets preparation.

\ifCLASSOPTIONcaptionsoff
  \newpage
\fi



%

\bibliographystyle{IEEEtran}

\bibliography{myrefsshort}

%

\clearpage

\end{document}